\documentclass[11pt]{article}
\usepackage{acl}
\usepackage{times}
\usepackage{latexsym}
\usepackage[T1]{fontenc}
\usepackage[utf8]{inputenc}
\usepackage{microtype}
\usepackage{inconsolata}
\usepackage{amsmath,amssymb,mathtools}
\usepackage{booktabs,tabularx,multirow}
\usepackage{graphicx}
\usepackage{algorithm}
\usepackage{algorithmic}
\usepackage{xcolor}
\usepackage{enumitem}
\usepackage{pifont}
\usepackage{placeins}
\newcommand{\method}{\textsf{PORA}}
\newcommand{\preserve}{\textsc{Preserve}}

\newcommand{\spend}{\textsc{Spend}}
\newcommand{\stage}[1]{\raisebox{0.5pt}{\textcircled{\raisebox{-0.9pt}{\footnotesize #1}}}}
\title{PORA: Learning When More Reasoning Pays Off for LLM Agents}
\author{
  \textbf{Bo Yan}$^{1}$, 
  \textbf{Qi Zhang}$^{2}$, 
  \textbf{Zhen Tan}$^{3}$, 
  \textbf{Tianlong Chen}$^{4}$, 
  \textbf{Song Wang}$^{1}$\thanks{~Corresponding author.} \\
  $^1$University of Central Florida \quad
   $^2$Independent Researcher \\
  $^3$Stevens Institute of Technology \quad
  $^4$UNC Chapel Hill \\
  \texttt{bo.yan@ucf.edu}, 
  \texttt{owl.zqq@gmail.com}, 
  \texttt{ztan21@stevens.edu} \\
  \texttt{tianlong@cs.unc.edu}, 
  \texttt{song.wang@ucf.edu}
}
\begin{document}
\maketitle
\begin{abstract}

LLM agents perform tasks over multiple steps and revise their actions based on feedback. As the total reasoning budget is often limited and must support future steps, the agent must choose between short and long reasoning before each action.
Longer reasoning may improve the final task outcome by preventing later errors, but it also uses budget capacity that may be needed for future decisions. 
This creates two challenges: estimating the downstream value of long reasoning and deciding when that value is worth using the remaining budget. 
We propose Paired Outcome-guided Reasoning Allocation (\method{}), which learns this decision by comparing short and long reasoning from the same interaction state. During training, \method{} completes the task twice from each selected state, once with a short reasoning allowance and once with a long allowance, and compares their final quality and total processing to estimate when longer reasoning is valuable. During inference, a controller weighs this estimate against learned prices on remaining-budget pressure and chooses either short or long reasoning without running both alternatives.
We evaluate \method{} across application, memory, code, retail, and mathematics tasks, achieving strong results across all settings. On BFCL memory, it exceeds the strongest existing method by 17.4--32.9 percentage points. Within the same agent, it matches or exceeds always using long reasoning on code and mathematics with 50.6\% fewer actor tokens and 31.2\% fewer model requests, respectively. 
%

\end{abstract}

\section{Introduction}
\label{sec:intro}

\begin{figure}[t]
\centering
\includegraphics[width=\linewidth]{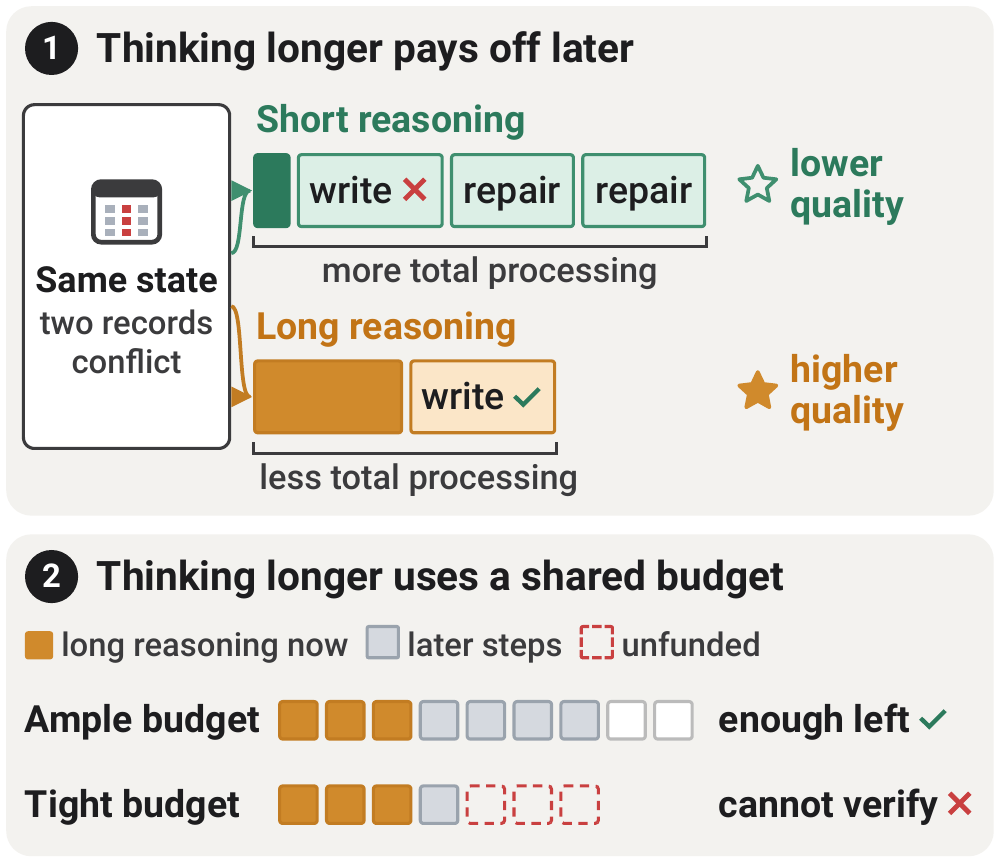}
\caption{\textbf{Deciding how long to reason at one step} (illustrative). \ding{182}~From the same state, short reasoning leads to a wrong write and two repairs, while long reasoning writes correctly; which is better is known only after the task ends. \ding{183}~Long reasoning uses budget that later steps also need, so the same choice can help with ample budget and hurt with a tight one.}
\label{fig:teaser}
\end{figure}

LLM agents solve tasks by using tools, maintaining memory, and revising their actions in response to execution feedback \citep{yao2023react,shinn2023reflexion,trivedi2024appworld,bfcl2026leaderboardsnapshot}. Under a finite episode budget, an agent must balance how much computation to allocate before each action. Specifically, a short reasoning allowance saves computation for later steps, while a longer allowance may help the agent resolve uncertainty before acting \citep{wei2022chain,snell2025scaling}. For example, when updating a calendar with conflicting records, additional reasoning may prevent an incorrect write and the repairs that follow (Figure~\ref{fig:teaser}). However, the extra computation also consumes a finite episode budget that may be needed for later evidence acquisition and verification. The value of reasoning therefore depends on its effect on the rest of the task, not only on the current action. This raises an important question: \emph{given the current interaction state and the remaining budget, when is it worth using a longer reasoning allowance?}


Since long reasoning is often wasted on easy inputs \citep{chen2025overthinking,sui2025stop}, existing methods choose reasoning budgets based on task complexity, plans, or confidence \citep{fang2025thinkless,han2025tokenbudgetaware,lin2026plan,fu2025certaindex,lee2026dart}. For agents, ARES estimates the effort needed for successful step completion \citep{yang2026ares}, and TAB learns turn-level budgets from trajectory rewards under a global token constraint \citep{jali2026turns}. Re-FORC estimates the reward available from additional thinking \citep{zabounidis2026reforc}.
These signals describe step difficulty, the payoff of further thinking, or the return of a whole trajectory, but none measures how the rest of an interaction changes when only the current allowance changes, with the interaction state and the subsequent policy held fixed.
Closing this gap raises two challenges.
\ding{182} \textbf{Quantifying the Benefit of Long Reasoning.}
A longer allowance at the current state can change subsequent actions and observations. It may improve terminal quality by preventing downstream errors, or reduce total processing by avoiding later corrective work. Local step success and current-step token counts therefore cannot supervise this value, which has to be measured through the completed interaction.
\ding{183} \textbf{Choosing Long Reasoning Under a Shared Budget.}
Even when longer reasoning helps, it uses capacity that later steps may need. The  interaction state can therefore warrant different allowances depending on the remaining budget and the number of future decision rounds. An allocation rule must weigh the expected benefit of longer reasoning against this capacity pressure.

We propose \emph{Paired Outcome-guided Reasoning Allocation} (\method{}), which learns when to use short or long reasoning by comparing their outcomes from the same interaction state. To address \ding{182}, during training we start two separate task executions from each selected state: one uses a short reasoning allowance (\preserve{}), and the other uses a long allowance (\spend{}). Both executions follow the same subsequent policy until the task ends, and we record their final quality and total processing. A small controller learns to predict the paired differences from the visible interaction history.
To address \ding{183}, the controller turns the remaining budget and future decision rounds into two capacity-pressure terms and weighs them, with learned positive prices, against the predicted benefit. At inference, it makes a single choice at each step and does not run both alternatives. Pairs can be recollected under the updated controller, since it determines how preserved capacity is later used.
Overall, \method{} enables an interactive agent to spend more reasoning when it is useful and save capacity when it is not.

We evaluate \method{} across application, memory, code, retail, and mathematics tasks, achieving strong results across all settings. On three BFCL memory settings, \method{} exceeds the strongest existing method by 17.4--32.9 percentage points. Within the same agent, \method{} improves over always using the short allowance in all six settings with complete fixed-allowance runs, and on code and mathematics it matches or exceeds always using the long allowance while using 50.6\% fewer actor tokens and 31.2\% fewer model requests, respectively. Refitting the controller with targets shuffled across states lowers code success by 7.9 points, showing the value of learning from paired outcomes.
Our contributions are as follows:
\begin{itemize}[leftmargin=*,itemsep=1pt,topsep=2pt]
\item \textbf{Problem.} We formulate reasoning allocation in interactive LLM agents as a same-state comparison: a longer allowance is valued by its effect on final quality and later processing, under the same subsequent policy.

\item \textbf{Method.} We propose \method{}, which learns this value from paired executions and selects the allowance by weighing the predicted benefit against learned prices on remaining budget.

\item \textbf{Experiments.} We evaluate \method{} across application, memory, code, retail, and mathematics tasks, showing that it improves task success and processing efficiency over existing methods and fixed allowances, and we analyze the value of paired outcomes and capacity prices.
\end{itemize}

\section{Methodology}
\label{sec:method}

\subsection{Problem Formulation}
\label{sec:problem}

An interactive LLM agent reasons before taking each task action. At decision $t$, a controller observes the visible history $s_t$, the remaining episode budget $b_t$ out of a total bank $B$, and the remaining decision rounds $N_t$. It chooses between a short reasoning allowance $\kappa_s$ (\preserve{}) and a long allowance $\kappa_l>\kappa_s$ (\spend{}). An allocation policy $\rho$ maps these inputs to a binary proposal and its allowance:
\begin{equation}
\begin{aligned}
\widetilde a_t &= \rho(s_t,b_t,N_t)\in\{0,1\},\\
\widetilde\kappa_t &= (1-\widetilde a_t)\kappa_s+\widetilde a_t\kappa_l.
\end{aligned}
\label{eq:allocation-problem}
\end{equation}
Here $0$ denotes \preserve{} and $1$ denotes \spend{}. Native phase and budget rules $\mathcal F$ map the proposal to a feasible grant $\kappa_t=\mathcal F(\widetilde a_t,s_t,b_t,N_t)$, which may differ from $\widetilde\kappa_t$. A fixed LLM actor acts under $\kappa_t$, the observation extends the history, and $b_{t+1}=b_t-c_t$, where the charge $c_t$ is the granted allowance or the realized use, depending on the interface.

Our goal is to allocate reasoning where it improves the completed task while preserving capacity for later steps. The task's evaluator supplies terminal quality $Y$, and we measure complete processing $C$, in a declared unit such as tokens, from the current decision to termination. $C$ includes subsequent work such as retrievals, repairs, and failed requests, differing from the charges. The controller uses only information available before allocation.

\subsection{Paired Outcome-guided Reasoning Allocation}
\label{sec:pora}

\begin{figure*}[t]
\centering
\includegraphics[width=\textwidth]{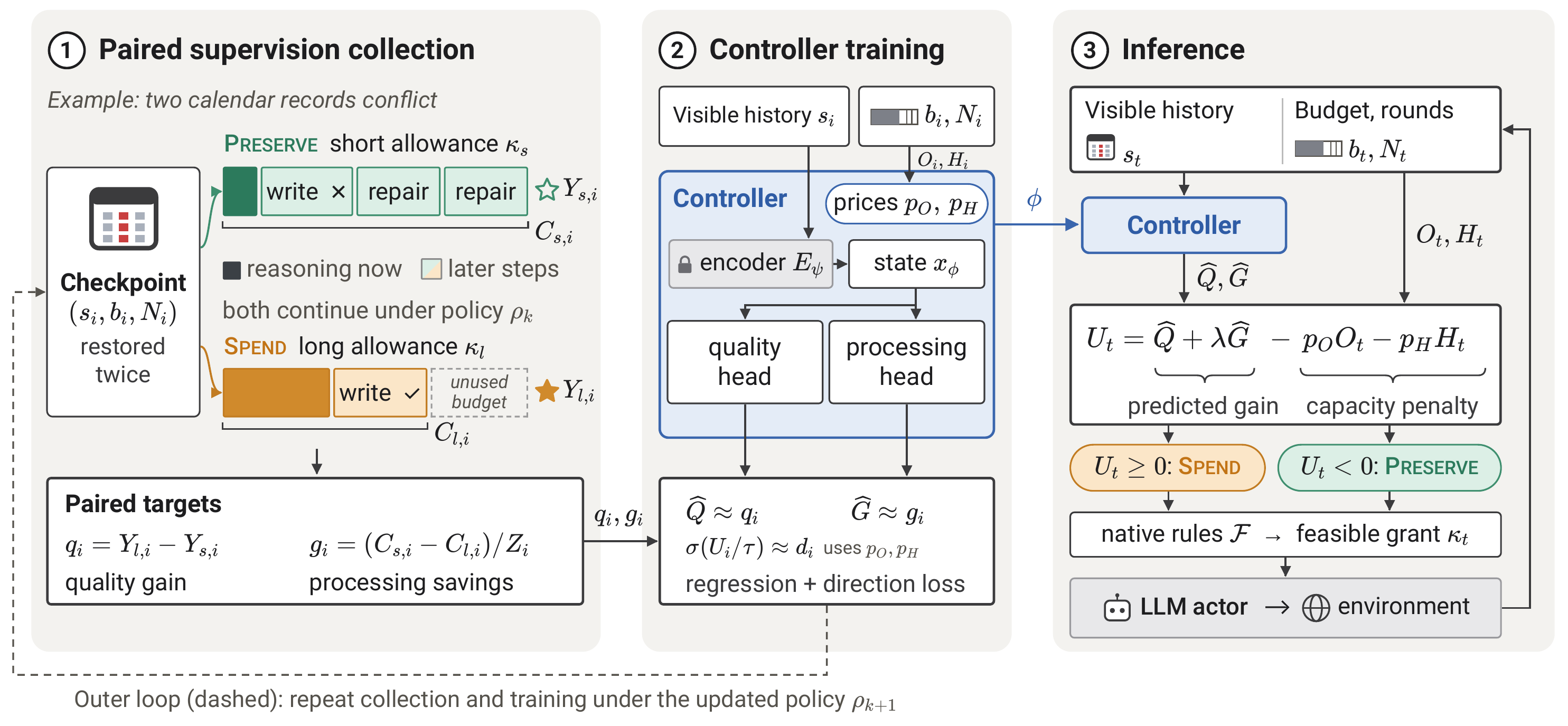}
\vspace{-12pt}
\caption{\textbf{Overview of \method{}.} \stage{1}~From one checkpoint $(s_i,b_i,N_i)$, a \preserve{} branch and a \spend{} branch continue under the same policy $\rho_k$; their differences in terminal quality and complete processing give the paired targets $(q_i,g_i)$. \stage{2}~The controller learns to predict these targets from the visible history and learns the prices on the pressure terms $(O_i,H_i)$ through the direction label $d_i$. \stage{3}~At inference, the predicted gain is weighed against the priced pressure, and native rules map the proposal to a feasible grant $\kappa_t$. The dashed arrow repeats collection and training under the updated policy. The calendar case is illustrative.}
\label{fig:method}
\vspace{-12pt}
\end{figure*}

\paragraph{Overview.}
Figure~\ref{fig:method} shows \method{} in three stages. Paired rollouts (\stage{1}) measure how short and long allowances differ in terminal quality and complete processing, addressing \ding{182}. A controller (\stage{2}) learns to predict these differences and to price remaining-budget pressure. At inference (\stage{3}), it weighs its predictions against the priced pressure, addressing \ding{183}. Collection and training can be repeated as the policy changes.

\paragraph{Paired Supervision Collection.}
\label{sec:paired-futures}

The benefit of additional reasoning may appear several steps later. In the calendar example, a short allowance may leave a record conflict that causes later repairs, while a long allowance may resolve it before a write. We thus construct supervision from completed trajectories rather than local step success.

At training iteration $k$, we collect decision checkpoints under an allocation policy $\rho_k$, starting from an initial policy $\rho_0$. For checkpoint $i$, we restore the same history, environment, memory, budget, and random state into two isolated branches. One receives the short allowance (\preserve{}), and the other receives the long allowance (\spend{}). Both then follow $\rho_k$ until termination, so the branches differ only in the allowance at the checkpoint, and their later actions and observations can diverge as a consequence. The memory instances draw their checkpoints from synthetic fixtures under the native memory interfaces (Appendix~\ref{sec:fit-recipes}).

Let $(Y_{s,i},C_{s,i})$ and $(Y_{l,i},C_{l,i})$ be the terminal quality and complete processing of the short and long branches. They provide two regression labels and a training set:
\begin{equation}
\begin{aligned}
q_i &= Y_{l,i}-Y_{s,i},\\
g_i &= (C_{s,i}-C_{l,i})/Z_i,\\
\mathcal D_k &= \bigl\{((s_i,b_i,N_i),(q_i,g_i))\bigr\}_{i=1}^{M_k}.
\end{aligned}
\label{eq:paired-target}
\end{equation}
$M_k$ denotes the number of pairs and $Z_i=\max(C_{s,i},C_{l,i},z_{\min})$, with $z_{\min}>0$, normalizes processing. Positive $q_i$ means that long allowance improves terminal quality; positive $g_i$ means that it saves complete processing despite allowing more reasoning now. For a tradeoff $\lambda\geq0$, $q_i+\lambda g_i$ equals the difference between the branch utilities $Y-\lambda C/Z_i$ of the long and short branches. Each branch runs once, so $(q_i,g_i)$ is a single-sample estimate of the expected paired difference under $\rho_k$, whose conditional mean is estimated below.

\paragraph{Controller Training.}
\label{sec:values}

The controller needs to recognize relevant evidence and unfinished work from the visible history. Let $u^{\mathrm{task}}$ be the task text and $u^{\mathrm{plan}}$ a textual evidence plan describing the evidence the task requires; both are visible from the first decision. A frozen pretrained encoder $E_\psi$ embeds these texts and the observations already returned by tools. We construct a compact controller representation as
\begin{equation}
\begin{aligned}
e^{\mathrm{task}} &= \operatorname{norm}\!\left(E_\psi(u^{\mathrm{task}})\right),\\
e^{\mathrm{plan}} &= \operatorname{norm}\!\left(E_\psi(u^{\mathrm{plan}})\right),\\
x_\phi(s) &= [r(s);\,f(s);\\
&\qquad \operatorname{proj}_\phi(e^{\mathrm{task}},e^{\mathrm{plan}});\,\omega(s)].
\end{aligned}
\label{eq:state-contract}
\end{equation}
Here $\operatorname{norm}$ denotes L2 normalization, also applied to observation embeddings, and the brackets denote concatenation. The features $r(s)$ summarize cosine similarities between the task, plan, and accumulated or recent observations, capturing evidence relevance. The feedback features $f(s)$ summarize interaction progress and tool successes or failures. The function $\operatorname{proj}_\phi$ concatenates learned affine projections of the task and plan embeddings, providing semantic context beyond their similarity summaries. Finally, $\omega(s)$ records visible obligations, such as missing evidence, pending writes, or repair needs. The budget quantities do not enter $x_\phi(s)$; they affect the decision only through the capacity terms in Eq.~\ref{eq:priced-exchange}. Exact features, dimensions, and encoder configurations are specified in Appendix~\ref{sec:feature-recipe} and the native recipes.

Two independent MLP heads read $x_i=x_\phi(s_i)$ and predict the paired consequences:
\begin{equation}
\begin{aligned}
\widehat Q_i &= \widehat Q_\phi(x_i)
=\operatorname{MLP}_Q(x_i;\phi_Q),\\
\widehat G_i &= \widehat G_\phi(x_i)
=\operatorname{MLP}_G(x_i;\phi_G).
\end{aligned}
\label{eq:value-predictions}
\end{equation}
The quality head estimates the terminal-quality gain $q_i$, and the processing head estimates normalized complete-processing savings $g_i$, both for the long allowance relative to the short one. Their parameter blocks $\phi_Q$ and $\phi_G$ are independent. Separate outputs keep quality and processing interpretable and allow an explicit tradeoff when choosing an allowance. The trainable parameters $\phi$ comprise the semantic projections, the two heads, and the capacity prices introduced below; through Eq.~\ref{eq:allocation-rule}, they define the allocation policy $\rho_\phi$.

To train both consequence estimates and allowance selection, we combine two regression losses with an allocation-direction loss. Let $U_i=U_\phi(s_i,b_i,N_i)$ be the capacity-priced allocation score of Eq.~\ref{eq:priced-exchange}, with pressure terms $O_i$ and $H_i$ computed from $(b_i,N_i)$. The shared per-example loss is
\begin{equation}
\begin{aligned}
\mathcal L_i(\phi)={}&w_Q\ell_Q(\widehat Q_i,q_i)
+w_G\ell_G(\widehat G_i,g_i)\\
&+w_D\ell_D(U_i/\tau,d_i).
\end{aligned}
\label{eq:learning-objective}
\end{equation}
We minimize the mean of these losses over $\mathcal D_k$. The first two terms fit the observed quality and processing differences. The third is binary cross-entropy with logits on the direction label $d_i=\mathbf{1}[q_i+\lambda g_i>\delta]$, which indicates whether the observed utility difference favors the long allowance by more than a margin $\delta\geq0$; $\tau>0$ scales the score, and the weights $w_Q,w_G,w_D$ control the loss contributions. The direction loss back-propagates through the projections, both heads, and the prices, and it is the only term that trains the prices. Since $d_i$ already reflects downstream budget effects under $\rho_k$, the prices act as a learned correction, fitted to the extent that $O_i$ and $H_i$ explain direction errors left by $\widehat Q_i+\lambda\widehat G_i$. Actor and encoder weights remain frozen. Appendix~\ref{sec:fit-recipes} specifies regression choices, label margins, tie handling, sample weighting, and additional task-specific loss terms.

After fitting, we select parameters $\phi_{k+1}$ with an instance-specific rule and set $\rho_{k+1}=\rho_{\phi_{k+1}}$. We can then collect new checkpoints and paired continuations on training tasks under $\rho_{k+1}$. This refresh matters because changing the continuation changes how preserved capacity is later used, and hence the labels to be predicted. Choosing \spend{} exactly when the expected paired utility difference under $\rho_k$ is nonnegative is the improvement step of rollout-based policy iteration \citep{lagoudakis2003approximate}; the controller approximates this step from visible inputs, and the capacity terms calibrate it to current budget pressure. Algorithm~\ref{alg:pora} and Appendix~\ref{sec:iteration-recipe} specify the loop; the application instance uses one recollection round.

\paragraph{Online Inference with Capacity Pricing.}
\label{sec:prices}

At inference, we encode the current visible history and construct $x_t=x_\phi(s_t)$ using the same representation as in training. Relative to a short allowance, the two heads estimate the long allowance's terminal-quality gain, $\widehat Q_\phi(x_t)$, and normalized complete-processing savings, $\widehat G_\phi(x_t)$. These are predictions of the paired consequences; inference does not execute either paired rollout.

A predicted benefit must also justify using capacity needed by later steps. In the calendar example, resolving a conflict now must leave budget for subsequent evidence acquisition and verification. We therefore derive two pressure terms in $[0,1]$ from the remaining budget and rounds (Eq.~\ref{eq:capacity-features}). $O_t$ measures the extra commitment $\kappa_l-\kappa_s$ relative to $b_t$, scaled by the share of rounds still to come. $H_t$ measures how far the budget left after a long grant falls short of funding short allowances in all future rounds, relative to $B$. Both vanish at the final decision. The allocation score is
\begin{equation}
\begin{aligned}
U_\phi(s_t,b_t,N_t)={}&\widehat Q_\phi(x_t)+\lambda\widehat G_\phi(x_t)\\
&-p_O O_t-p_H H_t.
\end{aligned}
\label{eq:priced-exchange}
\end{equation}
The first two terms combine predicted quality gain and processing savings using the tradeoff $\lambda$. The learned positive prices $p_O=\operatorname{softplus}(\alpha)$ and $p_H=\operatorname{softplus}(\eta)$ scale the capacity penalties.
Writing $U_t=U_\phi(s_t,b_t,N_t)$, the controller proposes
\begin{equation}
\widetilde a_t=
\begin{cases}
1\ (\spend{}), & U_t\geq0,\\
0\ (\preserve{}), & U_t<0.
\end{cases}
\label{eq:allocation-rule}
\end{equation}
The native rules then give the feasible grant $\kappa_t=\mathcal F(\widetilde a_t,s_t,b_t,N_t)$. The actor executes under $\kappa_t$. The returned observation and the charge $c_t$ update the inputs for the next decision. This cycle repeats until termination with all parameters fitted (details in Appendix~\ref{sec:capacity-recipe}).  

\definecolor{poragain}{HTML}{1F5FAE}
\definecolor{poramuted}{HTML}{7A7974}
\begin{table*}[t]
\centering\small
\setlength{\tabcolsep}{6.5pt}
\renewcommand{\arraystretch}{1.08}
\caption{\textbf{Task success rate (\%) on seven settings from five benchmarks.} Completed cells report percentages over the full task counts shown above. A dash marks a setting without a complete run. Fixed \preserve{} and fixed \spend{} run the \method{} agent with every allowance decision set to short or long; the LCB entries come from the shared-initial-program study, in which \method{} also solves 93 of 126 problems. Bold marks the highest completed rate in each column, including ties. The last row gives \method{} minus the best completed external method in percentage points. $^{\dagger}$Action-adapted configuration (Appendix~\ref{sec:comparison-recipes}).}
\label{tab:main-panel}
\begin{tabular}{@{}lccccccc@{}}
\toprule
& \textbf{AppWorld} & \multicolumn{3}{c}{\textbf{BFCL memory}} & \textbf{LCB} & \textbf{$\tau$-bench} & \textbf{MATH500} \\
\cmidrule(lr){3-5}
& & KV & Vector & Recursive & & retail & \\
\textcolor{poramuted}{\# tasks} & \textcolor{poramuted}{417} & \textcolor{poramuted}{155} & \textcolor{poramuted}{155} & \textcolor{poramuted}{155} & \textcolor{poramuted}{126} & \textcolor{poramuted}{40} & \textcolor{poramuted}{500} \\
\midrule
\multicolumn{8}{@{}l}{\textcolor{poramuted}{\textit{Reasoning-budget methods}}} \\
\quad Plan and Budget & 27.3 & 18.1 & 16.1 & 58.7 & 65.1 & 62.5 & 94.8 \\
\quad Certaindex & 4.1 & 41.9 & 15.5$^{\dagger}$ & 25.8$^{\dagger}$ & 39.7 & \textbf{72.5} & 73.4 \\
\addlinespace[3pt]
\multicolumn{8}{@{}l}{\textcolor{poramuted}{\textit{Search and rollout methods}}} \\
\quad DORA & 4.1 & 30.3 & 29.0 & 28.4 & 49.2 & 62.5 & 95.0$^{\dagger}$ \\
\quad BATS & 22.5 & 36.1 & 39.4 & 63.9 & 73.0 & 65.0 & 94.4$^{\dagger}$ \\
\quad LATS & 29/172 & 44.5 & 49.7 & 60.0 & 38.9 & 65.0 & \textbf{95.6}$^{\dagger}$ \\
\quad Forest-of-Thought & 7.0 & -- & -- & -- & 18.3 & \textbf{72.5} & 91.8 \\
\addlinespace[3pt]
\multicolumn{8}{@{}l}{\textcolor{poramuted}{\textit{\method{} agent with a fixed allowance}}} \\
\quad Fixed \preserve{} & -- & 74.2 & 71.0 & 79.4 & 68.3 & 70.0 & 91.8 \\
\quad Fixed \spend{} & -- & 73.5 & 74.8 & \textbf{83.9} & 73.0 & 70.0 & 95.0 \\
\midrule
\textbf{\textcolor{poragain}{\method{}}} & \textbf{30.9} & \textbf{77.4} & \textbf{81.3} & 81.3 & \textbf{73.8} & \textbf{72.5} & 95.4 \\
\quad \textcolor{poramuted}{$\Delta$ vs.\ best external} & \textcolor{poragain}{+3.6} & \textcolor{poragain}{+32.9} & \textcolor{poragain}{+31.6} & \textcolor{poragain}{+17.4} & \textcolor{poragain}{+0.8} & \textcolor{poramuted}{0.0} & \textcolor{poramuted}{$-$0.2} \\
\bottomrule
\end{tabular}
\end{table*}

\section{Experimental Setup}
\label{sec:experiments}

The experiments address three questions. \textbf{Q1}: how does \method{} compare with existing reasoning-budget and search methods in task success and processing cost? \textbf{Q2}: does state-dependent allocation improve over granting the same allowance at every step? \textbf{Q3}: how do the paired targets and the capacity prices contribute?

\paragraph{Benchmarks and metrics.}
We evaluate on five benchmarks covering seven settings. AppWorld \citep{trivedi2024appworld} provides 417 application tasks scored by \texttt{AppWorld.evaluate}. BFCL memory \citep{bfcl2026leaderboardsnapshot} provides 155 questions for each of three memory backends (key--value, vector, and recursive summary), scored by official validity. LiveCodeBench (LCB) \citep{jain2024livecodebench} provides 126 programming problems with public and private tests. The retail domain of $\tau$-bench \citep{yao2024taubench} provides 40 customer-service conversations scored by the task evaluator and its assertion judge. MATH500 \citep{hendrycks2021math} provides 500 problems with exact answer checking. We report the task success rate.

\paragraph{Models and budgets.}
The AppWorld actor is OpenAI gpt-oss-120b with a BGE-M3 encoder. All other settings use Google Gemma 4 31B IT with a MiniLM encoder. On AppWorld and BFCL, an episode has a bank of 8,192 reasoning tokens, a short allowance of 128, a long allowance of 1,024, and at most twenty rounds. The code, retail, and mathematics settings adapt the bank and allowances to their interfaces (Appendix~\ref{sec:quality-only}).

\paragraph{Controller training.}
Every setting uses the same controller architecture, with 2,400 to 4,960 trainable parameters, and fits its own weights. Paired rollouts come from public AppWorld training tasks, synthetic memory fixtures for BFCL, and earlier programming problems for LCB. The AppWorld and memory controllers are each fitted on no more than two dozen pairs; AppWorld uses one recollection round (Appendix~\ref{sec:fit-recipes}).

\paragraph{Baselines.}
Reasoning-budget methods decide how much the model reasons: Plan and Budget \citep{lin2026plan} and Certaindex \citep{fu2025certaindex}. Search and rollout methods spend additional rollouts, branches, or tool calls: DORA \citep{wang2025dora}, BATS \citep{liu2026bats}, LATS \citep{zhou2024lats}, and Forest-of-Thought \citep{bi2025forest}. Each baseline keeps its own procedure and runs with the same actor model inside the benchmark's tool interface (Appendix~\ref{sec:comparison-recipes}).

\paragraph{Controlled comparisons.}
For Q2, we run the \method{} agent with a fixed allowance: fixed \preserve{} always grants the short allowance, and fixed \spend{} grants the long allowance whenever the budget rules permit. These controls are complete for the three memory backends, LCB, $\tau$-bench retail, and MATH500; on LCB and MATH500, all policies start from the same initial programs or solutions, and the LCB rerun of \method{} from these programs again solves 93 of 126 problems. For Q3, we refit the controller with jointly shuffled targets, with absolute per-branch utilities in place of paired differences, with current-turn targets, and without the obligation features. We rescale the learned prices on vector memory, rescale the processing coefficient of a frozen MATH500 controller, and replay the 6,245 recorded AppWorld decision states with one term of Eq.~\ref{eq:priced-exchange} removed at a time.

\paragraph{Cost.}
Cost is the total number of model tokens, prompt plus completion, and the number of model requests. AppWorld totals count actor and encoder tokens on evaluated trajectories, and memory totals include memory preparation. Some recorded totals are lower bounds (Appendix~\ref{sec:resource-recipe}).

\begin{figure*}[t]
\centering
\includegraphics[width=\textwidth]{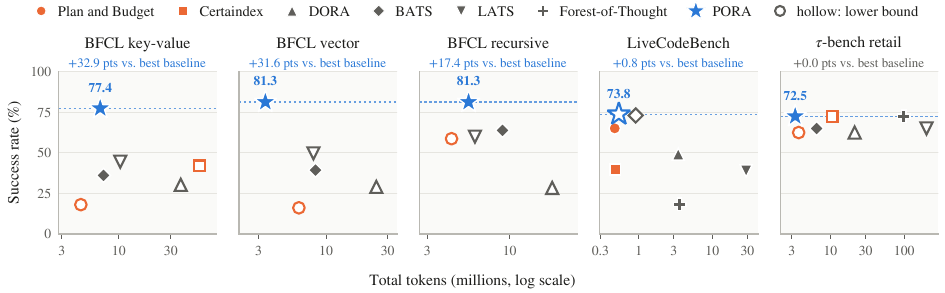}
\caption{\textbf{Task success against total tokens.} Each marker is one method with a recorded token total; hollow markers are lower bounds. The dotted line marks \method{}, and each subtitle gives its difference from the best external method in Table~\ref{tab:main-panel}. The logarithmic horizontal axis differs across panels.}
\label{fig:cost-quality}
\end{figure*}

\section{Results and Analysis}
\label{sec:results}

\subsection{Comparative Results (Q1)}
\label{sec:appworld-results}
\label{sec:memory-results}

\paragraph{Task success.}
In Table~\ref{tab:main-panel}, \method{} reaches the highest success rate on AppWorld, key--value memory, vector memory, and LCB, and shares the highest rate on $\tau$-bench retail. The margin is largest on BFCL memory, where \method{} reaches 77.4\%, 81.3\%, and 81.3\% on the three backends. These rates are 32.9, 31.6, and 17.4 points above the strongest external method in each setting. On AppWorld, \method{} completes 129 of 417 tasks against 114 for Plan and Budget, a 3.6-point difference. On LCB and MATH500, \method{} and the strongest external method are within one point of each other. MATH500 is close to saturation for this actor: even fixed \preserve{} solves 91.8\% of the problems. 

\paragraph{Processing cost.}
Figure~\ref{fig:cost-quality} plots success against total tokens for the five settings in which several methods have recorded totals. In all five, the recorded total of \method{} is below that of every search and rollout method. Plan and Budget records fewer tokens than \method{} on key--value memory, recursive-summary memory, and LCB, as does Certaindex on LCB, with lower success in each case. On $\tau$-bench retail, \method{}, Certaindex, and Forest-of-Thought reach 72.5\% with recorded totals of 3.3M, at least 10.5M, and 98.4M tokens. On AppWorld, where totals are recorded for \method{} and Plan and Budget, \method{} processes 67.3M actor and encoder tokens against 122.8M, a 45.2\% reduction.

\paragraph{Backbone generalization.}
Across additional LLM backbones evaluated on BFCL memory (Appendix~\ref{sec:backbone-study}), \method{} consistently sustains its advantage over fixed allocations, demonstrating that the learned capacity trade-offs generalize beyond a single base model.

\definecolor{poramuted}{HTML}{7A7974}
\begin{table}[t]
\centering\footnotesize
\setlength{\tabcolsep}{8.5pt}
\renewcommand{\arraystretch}{1.05}
\caption{\textbf{Controlled comparisons.} Each variant changes one factor of the \method{} run in its block header. $\Delta$ is \method{} minus the variant in points. Tokens are actor tokens in millions.}
\label{tab:ablation}
\begin{tabular}{@{}lrrr@{}}
\toprule
Variant & Succ. & $\Delta$ & Tokens \\
\midrule
\multicolumn{4}{@{}l}{\textcolor{poramuted}{\textit{Allocation policy}}} \\
\multicolumn{4}{@{}l}{BFCL vector (\method{}: 81.3, 3.39M tokens)} \\
\quad Fixed \preserve{} & 71.0 & 10.3 & 3.14 \\
\quad Fixed \spend{} & 74.8 & 6.5 & 3.84 \\
\multicolumn{4}{@{}l}{LCB, shared initial programs (\method{}: 73.8, 0.51M)} \\
\quad Fixed \preserve{} & 68.3 & 5.6 & 0.26 \\
\quad Fixed \spend{} & 73.0 & 0.8 & 1.03 \\
\multicolumn{4}{@{}l}{MATH500, shared initial solutions (\method{}: 95.4)} \\
\quad Fixed \preserve{} & 91.8 & 3.6 & $\geq$0.30 \\
\quad Fixed \spend{} & 95.0 & 0.4 & $\geq$0.89 \\
\midrule
\multicolumn{4}{@{}l}{\textcolor{poramuted}{\textit{Supervision target and state}}} \\
\multicolumn{4}{@{}l}{LCB, shared initial programs (\method{}: 73.8, 0.51M)} \\
\quad Shuffled targets & 65.9 & 7.9 & 0.29 \\
\quad Absolute utility & 73.0 & 0.8 & $\geq$0.53 \\
\multicolumn{4}{@{}l}{LCB, obligation refit (full state: 71.4, 0.51M)} \\
\quad Obligations removed & 65.1 & 6.3 & -- \\
\multicolumn{4}{@{}l}{BFCL recursive, 24-pair refit (terminal targets: 78.7)} \\
\quad Shuffled targets & 76.1 & 2.6 & 4.91 \\
\quad Current-turn targets & 77.4 & 1.3 & 5.76 \\
\midrule
\multicolumn{4}{@{}l}{\textcolor{poramuted}{\textit{Capacity prices}}} \\
\multicolumn{4}{@{}l}{BFCL vector (\method{}: 81.3)} \\
\quad Prices $\times 0.5$ & 72.9 & 8.4 & $\geq$3.52 \\
\quad Prices $\times 2$ & 76.1 & 5.2 & $\geq$3.40 \\
\bottomrule
\end{tabular}
\end{table}

\subsection{Effect of State-dependent Allocation (Q2)}
\label{sec:fixed-results}

\begin{figure}[t]
\centering
\includegraphics[width=\linewidth]{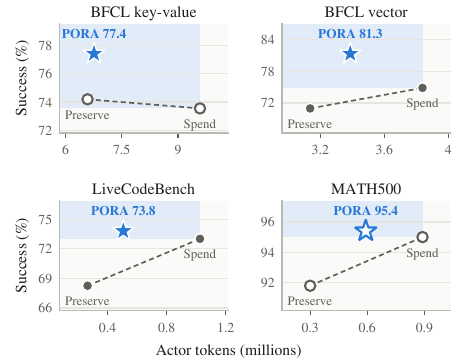}
\caption{\textbf{Fixed and learned allocation in the same agent} where all three token totals are recorded. Shading marks the region cheaper and more accurate than fixed \spend{}; hollow markers are lower bounds.}
\label{fig:fixed-policy}
\end{figure}

The fixed-allowance rows separate the allocation decision from the rest of the agent. Figure~\ref{fig:fixed-policy} places the three policies by success and actor tokens. On key--value memory, \method{} reaches 77.4\% with 6.76M tokens, against 73.5\% for fixed \spend{} with at least 9.59M. On vector memory, it is 10.3 points above fixed \preserve{} and 6.5 above fixed \spend{}, with a token total between the two. On LCB, it solves 93 of 126 problems against 86 and 92 and uses 50.6\% fewer actor tokens than fixed \spend{}. On MATH500, it solves 477 of 500 problems against 459 and 475 with 31.2\% fewer requests than fixed \spend{}. In these four settings, \method{} matches or exceeds fixed \spend{} at a lower recorded cost. Table~\ref{tab:ablation} summarizes these results.

On $\tau$-bench retail, the three policies complete 29, 28, and 28 of 40 tasks with nearly equal token totals (3.33M, 3.38M, and 3.30M), so the lower total relative to Forest-of-Thought is shared by the fixed-allowance agent.

\subsection{Effect of Paired Supervision and Capacity Pricing (Q3)}
\label{sec:mechanism-results}
\label{sec:readout-results}

\paragraph{Supervision target.}
On LCB, all supervision variants share the fitting problems, the initial programs, and the executor (Appendix~\ref{sec:matched-code}). As summarized in Table~\ref{tab:ablation}, jointly shuffling the quality and processing targets across states lowers the solved count from 93 to 83, a 7.9-point difference. Absolute per-branch utilities give 92. On recursive-summary memory, a 24-pair refit with shuffled or current-turn targets scores 2.6 and 1.3 points below the terminal-target refit. In a separate LCB refit, removing the obligation features lowers the solved count from 90 to 82, a 6.3-point difference.

\begin{table}[!t]
\centering\small
\caption{Native MATH500 sensitivity to the processing coefficient with a frozen controller.}
\label{tab:math-processing-sensitivity}
\begin{tabular}{lccc}
\toprule
Coefficient & Success & Repairs & Requests \\
\midrule
$0.5\times$ & 475/500 & 207 & 712 \\
$1.0\times$ & 476/500 & 187 & 689 \\
$2.0\times$ & 474/500 & 153 & 655 \\
\bottomrule
\end{tabular}
\end{table}

\paragraph{Capacity prices and processing coefficient.}
Scaling both learned prices by 0.5 or by 2 on vector memory lowers success from 81.3\% to 72.9\% and 76.1\% (Table~\ref{tab:ablation}). With a frozen MATH500 controller, scaling the processing coefficient by 0.5, 1, and 2 gives 475, 476, and 474 correct answers with 712, 689, and 655 requests (Table~\ref{tab:math-processing-sensitivity}). A larger coefficient lowers the number of requests while correctness stays within two problems, so the processing term mainly adjusts cost.


\section{Related Work}
\label{sec:related}

\paragraph{Sequential computation allocation.}
TAB treats multi-turn reasoning as a sequential budget problem, explicitly considers downstream dependencies and delayed feedback, and trains a budgeter with GRPO using accuracy and global token constraints \citep{jali2026turns}. Its trajectory-level reward supplies a shared advantage to the turn-level allocation decisions. ARES predicts the minimum suitable reasoning effort from interaction history using labels for successful step completion \citep{yang2026ares}. \method{} instead compares two allowances from the same state under a common continuation policy.

\paragraph{Predicting the value of reasoning.}
Re-FORC predicts expected future reward as a function of additional thinking tokens and uses it for stopping and resource selection \citep{zabounidis2026reforc}. \method{} instead predicts how two allowances change the rest of an interaction. Thinkless learns to select short or long reasoning, and Token-Budget-Aware Reasoning adjusts budgets according to problem complexity \citep{fang2025thinkless,han2025tokenbudgetaware}. Budget forcing and length-controlled reinforcement learning set the length of a single reasoning trace \citep{muennighoff2025s1,aggarwal2025l1}. Certaindex uses confidence signals to adapt computation, Plan and Budget allocates effort through a plan, and DART uses draft agreement and entropy to route thinking \citep{fu2025certaindex,lin2026plan,lee2026dart}. \method{} also prices the remaining episode budget.

\paragraph{Tool use and search.}
ReAct interleaves reasoning with actions \citep{yao2023react}; AppWorld and $\tau$-bench evaluate longer tool-mediated tasks \citep{trivedi2024appworld,yao2024taubench}. LATS, DORA, BATS, and Forest-of-Thought allocate search, rollouts, or tool effort \citep{zhou2024lats,wang2025dora,liu2026bats,bi2025forest}. We compare with them as complete systems.

\paragraph{Metareasoning.}
Valuing computation by its expected downstream benefit has a long history \citep{russell1991metareasoning,hay2012selecting,callaway2018learning}. Rational Metareasoning for LLMs uses a value-of-computation reward with Expert Iteration to learn selective reasoning \citep{desabbata2025rational}. Rollout-based approximate policy iteration evaluates alternative actions from sampled states under a common continuation policy and learns an updated policy \citep{lagoudakis2003approximate}. \method{} applies these principles to reasoning allowances in interactive tasks.

\section{Conclusion}
\label{sec:conclusion}

We presented \method{}, which learns when an interactive LLM agent should use short or long reasoning by comparing the two allowances from the same interaction state. Paired executions measure how a longer allowance changes final quality and total processing, a compact controller learns to predict these differences, and learned prices on remaining-budget pressure turn the predictions into a feasible allowance. Pairs are recollected as the controller changes, since it determines how preserved capacity is later used. Across five benchmarks, \method{} achieves strong task success; within the same agent, it matches or exceeds always using long reasoning on code and mathematics at lower cost, and shuffled paired targets lower code success.

\section*{Limitations}
\label{sec:limitations}

Each task family uses its own fitted controller and native phase and budget rules, which are specified per interface. Paired collection requires restoring the environment, memory, and budget into two isolated branches and running both to termination, which adds training cost and is not available for every environment; the memory instances therefore draw their checkpoints from synthetic fixtures. Each branch is executed once, and some controllers are fitted on a few dozen pairs or fewer, so the paired targets are noisy estimates. The controller chooses between two allowances and does not consider finer-grained budgets. Most results use one actor per setting; with the additional actors in Appendix~\ref{sec:backbone-study}, the differences between allocation policies are smaller.

\section*{Use of AI Assistants}
We used AI assistants to help edit the text. The authors designed the method and experiments, checked all content and results, and take full responsibility for the paper.

\section*{Ethics Statement}
The study uses benchmark-provided environments and licensed task data. All reported actions remain inside the permissions exposed by each benchmark. Deployment in consequential settings requires task-specific authorization and monitoring.
\section*{Reproducibility Statement}
The appendix defines the state contract, paired targets, model heads, capacity prices, native feasibility interface, selected populations, scoring boundaries, and resource units.
\bibliography{references}

@article{russell1991metareasoning,
  title = {Principles of Metareasoning},
  author = {Russell, Stuart and Wefald, Eric},
  journal = {Artificial Intelligence},
  volume = {49},
  number = {1--3},
  pages = {361--395},
  year = {1991}
}

@misc{hay2012selecting,
  title = {Selecting Computations: Theory and Applications},
  author = {Hay, Nicholas and Russell, Stuart and Tolpin, David and Shimony, Solomon E.},
  year = {2012},
  eprint = {1207.5879},
  archivePrefix = {arXiv}
}

@misc{callaway2018learning,
  title = {Learning to Select Computations},
  author = {Callaway, Frederick and Rangel, Antonio and Griffiths, Thomas L.},
  year = {2018},
  eprint = {1711.06892},
  archivePrefix = {arXiv}
}

@misc{desabbata2025rational,
  title = {Rational Metareasoning for Large Language Models},
  author = {De Sabbata, C. Nicol{\`o} and Sumers, Theodore R. and AlKhamissi, Badr and Bosselut, Antoine and Griffiths, Thomas L.},
  year = {2025},
  eprint = {2410.05563},
  archivePrefix = {arXiv}
}

@misc{han2025tokenbudgetaware,
  title = {Token-Budget-Aware LLM Reasoning},
  author = {Han, Tingxu and Wang, Zhenting and Fang, Chunrong and Zhao, Shiyu and Ma, Shiqing and Chen, Zhenyu},
  year = {2025},
  eprint = {2412.18547},
  archivePrefix = {arXiv}
}

@inproceedings{fang2025thinkless,
  title = {Thinkless: LLM Learns When to Think},
  author = {Fang, Gongfan and Ma, Xinyin and Wang, Xinchao},
  booktitle = {Advances in Neural Information Processing Systems},
  year = {2025},
  eprint = {2505.13379},
  archivePrefix = {arXiv}
}

@inproceedings{zabounidis2026reforc,
  title = {Re-FORC: Adaptive Reward Prediction for Efficient Chain-of-Thought Reasoning},
  author = {Zabounidis, Renos and Golatkar, Aditya and Kleinman, Michael and Achille, Alessandro and Xia, Wei and Soatto, Stefano},
  booktitle = {Proceedings of the 43rd International Conference on Machine Learning},
  volume = {306},
  series = {Proceedings of Machine Learning Research},
  year = {2026},
  eprint = {2511.02130},
  archivePrefix = {arXiv}
}

@inproceedings{fu2025certaindex,
  title = {Efficiently Scaling {LLM} Reasoning with Certaindex},
  author = {Fu, Yichao and Chen, Junda and Zhu, Siqi and Fu, Zheyu and Dai, Zhongdongming and Zhuang, Yonghao and Ma, Yian and Qiao, Aurick and Rosing, Tajana and Stoica, Ion and Zhang, Hao},
  booktitle = {Advances in Neural Information Processing Systems},
  year = {2025},
  eprint = {2412.20993},
  archivePrefix = {arXiv}
}

@inproceedings{lin2026plan,
  title = {Plan and Budget: Effective and Efficient Test-Time Scaling on Large Language Model Reasoning},
  author = {Lin, Junhong and Zeng, Xinyue and Zhu, Jie and Wang, Song and Shun, Julian and Wu, Jun and Zhou, Dawei},
  booktitle = {International Conference on Learning Representations},
  year = {2026},
  eprint = {2505.16122},
  archivePrefix = {arXiv}
}

@inproceedings{wang2025dora,
  title = {Every Rollout Counts: Optimal Resource Allocation for Efficient Test-Time Scaling},
  author = {Wang, Xinglin and Li, Yiwei and Feng, Shaoxiong and Yuan, Peiwen and Zhang, Yueqi and Shi, Jiayi and Tan, Chuyi and Pan, Boyuan and Hu, Yao and Li, Kan},
  booktitle = {Advances in Neural Information Processing Systems},
  year = {2025},
  eprint = {2506.15707},
  archivePrefix = {arXiv}
}

@inproceedings{liu2026bats,
  title = {Budget-Aware Tool Use Enables Effective Agent Scaling},
  author = {Liu, Tengxiao and Wang, Zifeng and Miao, Jin and Hsu, I-Hung and Yan, Jun and Chen, Jiefeng and Han, Rujun and Xu, Fangyuan and Chen, Yanfei and Jiang, Ke and Daruki, Samira and Liang, Yi and Wang, William Yang and Pfister, Tomas and Lee, Chen-Yu},
  booktitle = {Conference on Language Modeling},
  year = {2026},
  eprint = {2511.17006},
  archivePrefix = {arXiv}
}

@misc{jali2026turns,
  title = {Not All Turns Are Equally Hard: Adaptive Thinking Budgets for Efficient Multi-Turn Reasoning in Agents},
  author = {Jali, Neharika and Nayak, Anupam and Joshi, Gauri},
  year = {2026},
  eprint = {2604.05164},
  archivePrefix = {arXiv}
}

@misc{yang2026ares,
  title = {{ARES}: Adaptive Reasoning Effort Selection for Efficient {LLM} Agents},
  author = {Yang, Jingbo and Hou, Bairu and Wei, Wei and Bao, Yujia and Chang, Shiyu},
  year = {2026},
  eprint = {2603.07915},
  archivePrefix = {arXiv}
}

@inproceedings{yao2023react,
  title = {{ReAct}: Synergizing Reasoning and Acting in Language Models},
  author = {Yao, Shunyu and Zhao, Jeffrey and Yu, Dian and Du, Nan and Shafran, Izhak and Narasimhan, Karthik and Cao, Yuan},
  booktitle = {International Conference on Learning Representations},
  year = {2023}
}

@misc{yao2024taubench,
  title = {{$\tau$-bench}: A Benchmark for Tool-Agent-User Interaction in Real-World Domains},
  author = {Yao, Shunyu and Shinn, Noah and Razavi, Pedram and Narasimhan, Karthik},
  year = {2024},
  eprint = {2406.12045},
  archivePrefix = {arXiv}
}

@inproceedings{trivedi2024appworld,
  title = {{AppWorld}: A Controllable World of Apps and People for Benchmarking Interactive Coding Agents},
  author = {Trivedi, Harsh and Khot, Tushar and Hartmann, Mareike and Manku, Ruskin and Dong, Vinty and Li, Edward and Gupta, Shashank and Sabharwal, Ashish and Balasubramanian, Niranjan},
  booktitle = {Proceedings of the 62nd Annual Meeting of the Association for Computational Linguistics},
  pages = {16022--16076},
  year = {2024},
  eprint = {2407.18901},
  archivePrefix = {arXiv}
}

@misc{hendrycks2021math,
  title = {Measuring Mathematical Problem Solving with the {MATH} Dataset},
  author = {Hendrycks, Dan and Burns, Collin and Kadavath, Saurav and Arora, Akul and Basart, Steven and Tang, Eric and Song, Dawn and Steinhardt, Jacob},
  year = {2021},
  eprint = {2103.03874},
  archivePrefix = {arXiv}
}

@misc{jain2024livecodebench,
  title = {{LiveCodeBench}: Holistic and Contamination Free Evaluation of Large Language Models for Code},
  author = {Jain, Naman and Han, King and Gu, Alex and Li, Wen-Ding and Yan, Fanjia and Zhang, Tianjun and Wang, Sida and Solar-Lezama, Armando and Sen, Koushik and Stoica, Ion},
  year = {2024},
  eprint = {2403.07974},
  archivePrefix = {arXiv}
}

@inproceedings{bi2025forest,
  title = {Forest-of-Thought: Scaling Test-Time Compute for Enhancing {LLM} Reasoning},
  author = {Bi, Zhenni and Han, Kai and Liu, Chuanjian and Tang, Yehui and Wang, Yunhe},
  booktitle = {Proceedings of the 42nd International Conference on Machine Learning},
  pages = {4253--4267},
  volume = {267},
  series = {Proceedings of Machine Learning Research},
  publisher = {PMLR},
  year = {2025},
  url = {https://proceedings.mlr.press/v267/bi25a.html}
}

@inproceedings{zhou2024lats,
  title = {Language Agent Tree Search Unifies Reasoning, Acting, and Planning in Language Models},
  author = {Zhou, Andy and Yan, Kai and Shlapentokh-Rothman, Michal and Wang, Haohan and Wang, Yu-Xiong},
  booktitle = {Proceedings of the 41st International Conference on Machine Learning},
  pages = {62138--62160},
  volume = {235},
  series = {Proceedings of Machine Learning Research},
  publisher = {PMLR},
  year = {2024},
  url = {https://proceedings.mlr.press/v235/zhou24r.html}
}

@inproceedings{lee2026dart,
  title = {{DART}: Draft-Agreement Routing for Training-Free Adaptive Thinking Budgets in Hybrid Reasoning Models},
  author = {Lee, Jungseob and Hong, Seongtae and Lee, Seungjun and Seo, Jaehyung and Son, Junyoung and Eo, Sugyeong and Park, Chanjun and Park, Chanhee and Moon, Hyeonseok and Lim, Heuiseok},
  booktitle = {Findings of the Association for Computational Linguistics: EMNLP},
  year = {2026},
  eprint = {2606.23181},
  archivePrefix = {arXiv},
  url = {https://arxiv.org/abs/2606.23181v4}
}

@misc{bfcl2026leaderboardsnapshot,
  author = {{Berkeley Function Calling Leaderboard}},
  title = {{BFCL V4 Published Leaderboard Snapshot}},
  year = {2026},
  howpublished = {\url{https://gorilla.cs.berkeley.edu/leaderboard.html}},
  note = {Accessed 3 October 2026; page update 12 April 2026; evaluation revision f7cf735}
}

@inproceedings{lagoudakis2003approximate,
  title = {Reinforcement Learning as Classification: Leveraging Modern Classifiers},
  author = {Lagoudakis, Michail G. and Parr, Ronald},
  booktitle = {Proceedings of the Twentieth International Conference on Machine Learning},
  year = {2003},
  url = {https://users.cs.duke.edu/~parr/icml03.pdf}
}

@inproceedings{wei2022chain,
  title = {Chain-of-Thought Prompting Elicits Reasoning in Large Language Models},
  author = {Wei, Jason and Wang, Xuezhi and Schuurmans, Dale and Bosma, Maarten and Ichter, Brian and Xia, Fei and Chi, Ed and Le, Quoc V. and Zhou, Denny},
  booktitle = {Advances in Neural Information Processing Systems},
  volume = {35},
  pages = {24824--24837},
  year = {2022}
}

@inproceedings{shinn2023reflexion,
  title = {Reflexion: Language Agents with Verbal Reinforcement Learning},
  author = {Shinn, Noah and Cassano, Federico and Gopinath, Ashwin and Narasimhan, Karthik and Yao, Shunyu},
  booktitle = {Advances in Neural Information Processing Systems},
  volume = {36},
  pages = {8634--8652},
  year = {2023}
}

@inproceedings{snell2025scaling,
  title = {Scaling {LLM} Test-Time Compute Optimally Can be More Effective than Scaling Parameters for Reasoning},
  author = {Snell, Charlie and Lee, Jaehoon and Xu, Kelvin and Kumar, Aviral},
  booktitle = {International Conference on Learning Representations},
  year = {2025}
}

@inproceedings{chen2025overthinking,
  title = {Do {NOT} Think That Much for 2+3=? {O}n the Overthinking of Long Reasoning Models},
  author = {Chen, Xingyu and Xu, Jiahao and Liang, Tian and He, Zhiwei and Pang, Jianhui and Yu, Dian and Song, Linfeng and Liu, Qiuzhi and Zhou, Mengfei and Zhang, Zhuosheng and Wang, Rui and Tu, Zhaopeng and Mi, Haitao and Yu, Dong},
  booktitle = {Proceedings of the 42nd International Conference on Machine Learning},
  series = {Proceedings of Machine Learning Research},
  volume = {267},
  pages = {9487--9499},
  publisher = {PMLR},
  year = {2025}
}

@inproceedings{muennighoff2025s1,
  title = {s1: Simple test-time scaling},
  author = {Muennighoff, Niklas and Yang, Zitong and Shi, Weijia and Li, Xiang Lisa and Fei-Fei, Li and Hajishirzi, Hannaneh and Zettlemoyer, Luke and Liang, Percy and Cand{\`e}s, Emmanuel and Hashimoto, Tatsunori},
  booktitle = {Proceedings of the 2025 Conference on Empirical Methods in Natural Language Processing},
  pages = {20275--20321},
  address = {Suzhou, China},
  publisher = {Association for Computational Linguistics},
  doi = {10.18653/v1/2025.emnlp-main.1025},
  year = {2025}
}

@inproceedings{aggarwal2025l1,
  title = {L1: Controlling How Long A Reasoning Model Thinks With Reinforcement Learning},
  author = {Aggarwal, Pranjal and Welleck, Sean},
  booktitle = {Conference on Language Modeling},
  year = {2025}
}

@misc{sui2025stop,
  title = {Stop Overthinking: A Survey on Efficient Reasoning for Large Language Models},
  author = {Sui, Yang and Chuang, Yu-Neng and Wang, Guanchu and Zhang, Jiamu and Zhang, Tianyi and Yuan, Jiayi and Liu, Hongyi and Wen, Andrew and Zhong, Shaochen and Zou, Na and Chen, Hanjie and Hu, Xia},
  year = {2025},
  eprint = {2503.16419},
  archivePrefix = {arXiv}
}

\clearpage
\newpage
\appendix
\section{Method Details and Task Recipes}
\label{sec:appendix}
\label{sec:recipes}

This appendix gives the details behind Section~\ref{sec:method}: the controller features and parameter counts, the capacity terms and feasible grants, the fitting recipes, the paired collection loop, and the native settings of each task family.

\subsection{Feature construction and parameter accounting}
\label{sec:feature-recipe}

The input is an event tensor of shape $B_{\mathrm{batch}}\times T\times D$, a visibility mask, and role/status identifiers. Row zero is the task and row one is the evidence plan. Both anchors are visible. The remaining visible events retain their chronological order. Each embedding is L2-normalized with denominator floor $10^{-12}$, and the encoder output is detached during controller fitting.

Observation events have role identifier 3. Status identifiers 2 and 3 indicate success and failure. The twelve signals, in order, are: task--plan cosine; maximum task--observation and plan--observation cosines; position-weighted task--observation and plan--observation means; task and plan cosines with the last observation; visible dynamic-event count divided by 20 and clipped to $[0,1]$; failed and successful observation fractions; and the last observation's failure and success indicators. Position weights are the event index plus one. Observation-dependent values are zero for an empty observation set. Right padding is masked.

Equation~\ref{eq:state-contract} groups these signals by their role. The seven cosine signals form $r(s)$, and the five progress/status signals form $f(s)$. Let $v=[r(s);f(s)]\in\mathbb R^{12}$. Separate affine maps project the normalized task and plan embeddings from dimension $D$ to two dimensions each; their concatenation is $\operatorname{proj}_\phi$. Six obligation values $\omega(s)$, clipped to $[0,1]$, complete the representation. Its exact instantiation is
\begin{equation}
\begin{aligned}
x_\phi(s)=[&v;\,P_{\mathrm{task}}e^{\mathrm{task}}+a_{\mathrm{task}};\\
&P_{\mathrm{plan}}e^{\mathrm{plan}}+a_{\mathrm{plan}};\,\omega(s)]
\in\mathbb R^{22}.
\end{aligned}
\label{eq:state-instantiation}
\end{equation}
The dimension is $12+2+2+6=22$. The two independently parameterized heads are
\begin{equation}
\begin{aligned}
\operatorname{Head}_h(x) &= \tanh\!\left(w_{h,2}^{\mathsf T}\operatorname{GELU}(z_h)+a_{h,2}\right),\\
z_h &= W_{h,1}\operatorname{LN}_h(x)+a_{h,1},\quad h\in\{Q,G\}.
\end{aligned}
\label{eq:dual-heads}
\end{equation}
Both heads use LayerNorm with $\epsilon=10^{-5}$, an affine map from 22 to 16, exact GELU, an affine map from 16 to one, and $\tanh$. Each head has $44+368+17=429$ parameters. The two bottlenecks have $4(D+1)$ parameters; adding the two heads and two prices gives $4(D+1)+860$, or 2,400 for $D=384$ and 4,960 for $D=1024$.

\subsection{Capacity arithmetic and action boundary}
\label{sec:capacity-recipe}

The controller receives six native ledger values encoded as $\log(1+\cdot)$: total bank $B$, used budget, remaining bank $b$, remaining rounds $N$, short allowance $\kappa_s$, and long allowance $\kappa_l$. They are decoded before use and enter only the pressure features, never the 22-dimensional representation $x_\phi$. With $F_t=[N_t-1]_+$ and $[u]_+=\max(u,0)$, the clipped pressure features are
\begin{equation}
\begin{aligned}
O_t &= \operatorname{clip}_{[0,1]}\!\left(
\frac{[\kappa_l-\kappa_s]_+}{\max(b_t,1)}
\frac{F_t}{\max(N_t,1)}\right),\\
H_t &= \operatorname{clip}_{[0,1]}\!\left(
\frac{[\kappa_sF_t-[b_t-\kappa_l]_+]_+}{\max(B,1)}\right).
\end{aligned}
\label{eq:capacity-features}
\end{equation}
Denominator floors of one native unit define the empty-bank boundary. Both features vanish at the final decision. Initial prices are 0.05, implemented by initializing the raw parameters to inverse-softplus(0.05). The utility uses $\lambda=0.05$.

For interface $j$, the native phase and allowance rule $\mathcal F_j$ maps the core proposal to a grant, and $\chi_j$ gives the local charge $c_t$:
\begin{equation}
\begin{aligned}
\widetilde a_t &= \mathbf{1}[U_\phi(s_t,b_t,N_t)\geq0],\\
\kappa_t &= \mathcal F_j(\widetilde a_t,s_t,b_t,N_t),\\
b_{t+1} &= b_t-c_t,\quad c_t=\chi_j(\kappa_t,\text{realized use}).
\end{aligned}
\label{eq:feasible-action}
\end{equation}
The charging rule distinguishes committed allowance from generated-token use. An exhausted legal bank terminates the corresponding action sequence. Native terminal scoring applies to the resulting task outcome. The following recipes specify each interface's phase rules and execution limits.

\subsection{Fitting recipes}
\label{sec:fit-recipes}

\paragraph{Application fitting.}
The initial collection has twelve paired checkpoints at steps zero and seven from six public-training tasks. Recollection visits six different tasks from the same scenario families and supplies ten decision-observable pairs. The selected updated checkpoint has two quality-informative pairs and eight quality ties. Both fits use 300 full-batch AdamW epochs, learning rate 0.001, weight decay 0.001, and gradient-norm clipping at one; the update initializes from the earlier controller.

The application quality label is the Spend-minus-Preserve difference in terminal success plus 0.25 times the difference in requirement completion. Because both heads end in $\tanh$, labels whose magnitude exceeds one are fitted toward the saturated output. Processing is normalized by the maximum complete branch cost. The objective combines smooth-L1 quality loss, smooth-L1 processing loss, binary direction loss, and smooth-L1 utility loss with weights 1, 0.25, 0.25, and 0.25. Quality ties have quality-fit weight 0.25; their direction and utility paths detach the quality prediction so processing can receive the cost signal. Direction logits use temperature 0.25. These weights concentrate fitting on the paired quality contrast and the continuation-processing signal.

\paragraph{Memory fitting.}
Each measured memory instance uses 24 synthetic paired fixtures under its local native memory interface, with sixteen fitting and eight selection fixtures. Terminal correctness supplies $q$, and Eq.~\ref{eq:paired-target} supplies $g$ with a one-position floor. The shared loss components are introduced in Eq.~\ref{eq:learning-objective}; the measured memory instance uses the following batched objective:
\begin{equation}
\begin{aligned}
\mathcal L={}&\operatorname{MSE}(\widehat Q,q)
+\operatorname{MSE}(\widehat G,g)\\
&+0.25\operatorname{BCEWithLogits}(U_\phi/0.25,d),\\
d={}&\mathbf{1}[q+0.05g>0.02].
\end{aligned}
\label{eq:fit-loss}
\end{equation}
AdamW uses learning rate $3\times10^{-4}$ and weight decay $10^{-4}$ for at most 300 full-batch epochs. The selection score is the sum of quality and processing MSE on the eight selection fixtures; a recorded direction-first variant orders direction accuracy before that sum. The selected vector and recursive-summary fits use epochs 22 and 300. The support study on supervision uses a common 300-epoch refit on all 24 fixtures after its sixteen/eight development comparison. The support refit and main-table recursive-summary checkpoint retain separate source bindings and measured populations.

\subsection{Paired collection and policy refresh}
\label{sec:iteration-recipe}

Each checkpoint restores environment state, visible history, external memory, the budget ledger, and random state into isolated namespaces. Complete terminal scores and processing in one declared unit are required for both branches. The iteration loop keeps collection, fitting, and the instance's selection rule connected to an identified continuation policy. At deployment, Eq.~\ref{eq:feasible-action} applies the fitted controller through the native interface.

\begin{algorithm}[t]
\caption{Paired Outcome-guided Reasoning Allocation (\method{})}
\label{alg:pbpi}
\label{alg:pora}
\begin{algorithmic}[1]
\REQUIRE Initial continuation $\rho_0$, iteration count $K$.
\FOR{$k=0,\ldots,K-1$}
\STATE Collect checkpoints under $\rho_k$ on the designated collection tasks.
\STATE Restore each checkpoint into two isolated branches.
\STATE Grant $\kappa_s$ and $\kappa_l$; continue both under $\rho_k$ to termination.
\STATE Form visible states and paired targets using Eqs.~\ref{eq:paired-target} and~\ref{eq:state-contract}.
\STATE Fit the projections, value heads, and capacity prices on the fitting records.
\STATE Select $\rho_{k+1}$ using the instance's declared checkpoint rule.
\ENDFOR
\STATE Deploy $\rho_K$: predict values, price capacity, execute the native grant, and update the history and bank.
\end{algorithmic}
\end{algorithm}

\subsection{Application instance}
\label{sec:appworld-recipe}

\paragraph{State and encoder.}
AppWorld uses normalized 1,024-dimensional BGE-M3 embeddings. The six obligation slots contain required-source coverage, required-source count divided by eight and clipped at one, pending-write status, execution-failure fraction, last-execution failure, and successful-state-change status. Returned records update these fields before the next allowance.

\paragraph{Native execution.}
The bank is 8,192 committed reasoning tokens; short and long allowances are 128 and 1,024, with at most twenty rounds. Actor temperature is 0.6, top-$p$ is 0.95, context is 32,768, and each decision allows at most 2,048 additional answer tokens. The observation view takes the first 2,500 characters of each normalized response.

The phase becomes repair after an evidence-check rejection, two consecutive recovery-trigger events, or an acquisition with unchanged source coverage below $3/7$. A pending write, a successful state change, complete source coverage, or at most three remaining rounds selects commit. Other states acquire evidence. Six prior Spend decisions enforce later Preserve decisions. One eligible repair may receive recovery Spend; ordinary spending requires coverage at least $3/7$. Allowances are constrained by the remaining bank and charged in full.

The evidence-aware executor checks API names, fields, parameters, and the observed support of values. When a generated program has a valid read prefix, it can execute the longest qualifying top-level prefix to obtain evidence for a subsequent revision. The prefix contains an API call and passes the same support checks. AppWorld \texttt{0.1.3.post1} and \texttt{AppWorld.evaluate} define the final 417-task population and scorer.

\subsection{Memory instances}
\label{sec:memory-recipe}

\paragraph{Preparation and state.}
Each method completes native prerequisite conversations before answering the mode-specific 155 questions. Key--value memory uses explicit records; vector memory uses bounded topic updates and similarity retrieval; recursive summaries use the official 10,000-character cap. The controller encoder is MiniLM with attention-mask mean pooling, a 512-position tokenizer limit, and embedding dimension 384.

The two anchors are the question and the plan to recover the requested fact through native memory tools. The adapter includes the most recent eighteen visible events. Tool returns have observation role 3; other dynamic events have role 2. The six obligation fields represent observed successful evidence, a source-load value of $1/8$, a zero pending-write slot, tool-failure fraction, last-tool-failure status, and a zero state-change slot. These local fields feed the common graph in the same order.

\paragraph{Action and scorer.}
The native phase policy permits at most six ordinary Spend decisions and one eligible recovery Spend. Its source-coverage threshold is zero for the memory interface. The common bank is 8,192 with short and long allowances of 128 and 1,024 and at most twenty rounds. The core proposal is translated to the local short or long allowance. If the remaining bank cannot fund the long allowance, the executor attempts a complete short allowance; an insufficient short bank terminates. Granted allowances are charged in full. The official validity field scores every terminal answer. Shared-history scenario sizes are 30, 25, 25, 25, and 50.

\section{Comparisons and Statistical Detail}
\label{sec:comparison-recipes}

\subsection{External-method execution}

Plan and Budget uses state-dependent decomposition with 1,024 planner tokens in a 4,096 generation cap and a separate 8,192-token action bank on AppWorld. Its realized action thinking is charged and planning is recorded separately. \method{} commits granted allowances. Both use the SDK limit of 1,000 requests per code execution. BATS and Forest-of-Thought permit 100 new API requests across a search; DORA permits 100 per trajectory. DORA uses four rollout slots, three rounds, and 256-token increments. Forest-of-Thought uses three trees, branching two, sparse width one, depth three, and three votes. BATS uses budgeted tool use, verification, and selected-state restoration over at most twenty rounds.

Memory comparisons retain their declared planning, search, confidence, or rollout procedures inside the native memory interfaces. The complete additional Certaindex, DORA, and LATS rows use their native preparation and the same 155-question mode populations. Plan and Budget's complete recursive-summary run uses its released scheduling kernel and same-weight context recovery with 131,072-token context capacity. Its other three Gemma memory rows use the recorded thirty-step allowance. These execution envelopes define the system comparisons in Table~\ref{tab:main-panel}. LATS on AppWorld is evaluated on 172 of the 417 tasks. The action-adapted LATS run on MATH500 evaluates all 500 problems and obtains 478 correct answers. Fixed \spend{} on recursive-summary memory evaluates all 155 questions and obtains 130 correct answers.

The vector and recursive-summary Certaindex instances use an action-adapted confidence procedure: a 128-token reasoning cap, 64-token chunks, and two consecutive consistent answer probes. Each probe permits 256 tokens to express a native tool call; the upstream completion interface uses a 20-token probe. This configuration reaches its two-probe stopping check at the second and final chunk, using four physical requests per action and at most thirty action steps per turn. Both populations contain all 155 official outcomes, with 24 vector and 40 recursive-summary answers correct. Budget-exhausted trajectories receive native scoring within the same denominator. The retained-attempt ledger includes probes, recovery work, and native memory preparation.

On MATH500, the action-adapted DORA instance applies its released allocation kernel to three directions and four rollout allocations for one round. Candidate solutions have a 2,048-token cap and temperature 0.2; self-critique uses the visible problem and candidate, and the hidden answer is scored after selection with \texttt{verify\_MATH.grade\_answer}. The same Gemma actor revision and all 500 problem identities yield 475 correct answers, compared with 477 for \method{}. This complete endpoint comparison retains each method's recorded execution envelope.

\subsection{Paired inference and multiplicity}
\label{sec:stats}

The main-panel quality differences are AppWorld 3.60 points with grouped interval $[-0.48,7.67]$, key--value memory 32.90 with $[17.60,46.43]$, vector memory 31.61 with $[15.20,44.76]$, and recursive-summary memory 17.42 with $[4.67,28.75]$.

The bootstrap samples complete dependence groups with replacement and recomputes the task-weighted difference. AppWorld has 139 groups of three tasks; memory has five groups with sizes 30, 25, 25, 25, and 50. Every interval uses 20,000 draws. AppWorld quality uses seed 61006 and processing uses 61007. The main memory comparisons use seed 61008. The supervised-target and fixed-allocation controls use seed 20261006.

The four main quality comparisons have group sign-flip two-sided $p$ values 0.1041, 0.0625, 0.0625, and 0.2500 in table order; their Holm-adjusted values are all 0.25. The five-scenario exact test enumerates all 32 sign patterns, giving a minimum attainable two-sided $p$ of 0.0625. For AppWorld success, the task-level pairs comprise 54 wins, 39 losses, and 324 ties.

For the fixed-allocation family, raw $p$ values are 0.125 and 0.3125, with Holm values 0.25 and 0.3125. For the supervision family, raw $p$ values are 0.6875 and 0.875 and both Holm values are 1.0. The supervision rows yield task-level win/loss/tie counts of 11/7/137 and 13/11/131. The fixed-allocation rows yield 21/5/129 and 18/8/129. Scenario-level resampling preserves the shared preparation history.

\subsection{Readout sensitivity at recorded states}

The AppWorld study replays all 6,245 allocation states through the selected 4,960-parameter controller and the original native rules. Removing quality changes 2,951 core proposals and 1,002 final allowances. Removing processing changes 25 proposals and three final allowances. Removing prices changes twelve proposals and three final allowances; doubling prices changes five proposals and zero final allowances. Halving prices changes nine proposals and three final allowances. These measurements identify the contribution to recorded-state decisions at the selected operating point.

\subsection{Additional price and bank interventions}

The original support families also vary the price scale and shared bank on complete vector-memory populations. Table~\ref{tab:additional-controls} retains these comparisons beside their common scenario grouping. With the controller unchanged, a bank of 4,096 or 16,384 tokens gives 79.4\% and 75.5\%, against 81.3\% at the fitted bank of 8,192.

\begin{table}[t]
\centering\footnotesize
\caption{Additional vector-memory interventions. Differences are percentage points; adjusted $p$ uses the original price or budget family.}
\label{tab:additional-controls}
\begin{tabularx}{\linewidth}{@{}Xrrr@{}}
\toprule
Contrast & Gain & 95\% interval & Adj. $p$ \\
\midrule
Bank 16,384 vs 8,192 & $-5.81$ & $[-14.55,0.80]$ & $0.750$ \\
Bank 4,096 vs 8,192 & $-1.94$ & $[-5.00,0.00]$ & $0.750$ \\
Price $0.5\times$ vs $2\times$ & $-3.23$ & $[-11.11,3.00]$ & $0.750$ \\
Price $0.5\times$ vs $1\times$ & $-8.39$ & $[-14.29,-4.67]$ & $0.188$ \\
Price $1\times$ vs $2\times$ & $5.16$ & $[-0.80,9.50]$ & $0.750$ \\
\bottomrule
\end{tabularx}
\end{table}

\section{Resources and Additional Native Populations}
\label{sec:resource-recipe}

\subsection{Computing infrastructure}
\label{sec:compute}

All actors and encoders run on a shared cluster of NVIDIA DGX B200 nodes, each with eight B200 GPUs (180 GB memory per GPU). The controllers have 2,400 to 4,960 parameters and are fitted with full-batch AdamW. We report computation as the token and request counts given in this section rather than as GPU hours.

\subsection{Main-panel resources}

AppWorld's evaluated trajectories consume 64,844,347 actor positions and 2,443,846 encoder positions, totaling 67,288,193. Plan and Budget consumes 122,753,680 positions under the matched evaluated-trajectory scope. Dividing by 417 gives means of 161,363 and 294,373. The 45.18\% reduction has grouped interval $[41.87,48.28]\%$. Processing positions count tokenization-dependent model inputs and outputs.

\begin{table*}[t]
\centering\small
\caption{\method{} memory resources, including prerequisite preparation and retained attempts. Actor and encoder positions use their respective tokenizers. Service time sums request durations; task time sums retained task-attempt durations.}
\label{tab:memory-resources}
\begin{tabular}{@{}lrrrrr@{}}
\toprule
Memory mode & Actor positions & Request starts & Encoder positions & Service (s) & Task time (s) \\
\midrule
Key--value & 6,760,203 & 1,673 & 326,149 & 4,699.93 & 4,891.14 \\
Vector & 3,387,257 & 1,028 & 1,056,653 & 3,117.61 & 3,866.24 \\
Recursive summary & 5,483,736 & 1,431 & 409,287 & 15,604.28 & 15,917.76 \\
\bottomrule
\end{tabular}
\end{table*}

Vector encoder processing comprises 278,609 controller positions and 778,044 memory-backend positions.

\subsection{Native code, retail, and mathematics interfaces}
\label{sec:quality-only}

Table~\ref{tab:main-panel} includes the complete external quality comparisons for LiveCodeBench, $\tau$-bench retail, and MATH500. Their native interfaces apply the common controller to executable repair, customer and tool feedback, and exact answer checking. The recipes below define each action and scorer; the following subsections give the matched controls and resource comparisons.

\paragraph{Code.}
The adapter reads the problem, program, public test feedback, and remaining allowance, using 384-dimensional MiniLM features with a 256-position limit per field. Six obligation slots describe public correctness, public-test residual fraction, syntactic validity, repair phase, legal-cap fraction, and available feedback. Deployment starts with a 4,096-token program under an 8,192-token generated-token bank. At each of at most two repair decisions, the same predictor evaluates feasible caps of 512, 2,048, and 4,096 and Preserve with utility zero. The greatest utility selects the action; the smaller cap resolves ties. Generated tokens reduce the bank, and native code-generation metrics score the final program \citep{jain2024livecodebench}.

\method{} solves 93/126 and BATS solves 92/126, a difference of 0.79 points with contest-grouped interval $[-4.55,5.83]$ over 39 groups. The pairs contain seven wins, six losses, and 113 ties; grouped sign-flip $p=1.0$. The \method{} run records 179 request starts and at least 533,669 actor tokens.

\paragraph{Retail.}
The adapter reads the customer goal, retail policy, conversation prefix, tool feedback, and remaining bank. Its six slots describe successful evidence, pending calls, pending state changes, error rate, last error, and successful modifications. The native evaluator and registered Gemma assertion judge score the forty tasks. The complete selected \method{} outcome is 29/40, matching the strongest measured Certaindex and Forest-of-Thought endpoints. Resource collection includes actor, simulator, assertion judge, and retained attempts in its declared scope.

\paragraph{Mathematics.}
The adapter reads the problem, current draft, completion status, boxed-answer obligation, visible draft evidence, and remaining bank. Six obligations describe boxed-answer presence, cap termination, draft length, balanced braces, repair feasibility, and draft-token use. A 1,024-token draft uses a 2,560-token bank; Spend requests a feasible repair capped at 1,536 and Preserve accepts the draft. At this final repair, both future-capacity terms are zero. The fit combines quality and processing squared losses with 0.25 times binary direction loss, with weights 32, 8, and 0.25 for quality-improving, cost-sensitive, and tied strata. AdamW uses learning rate $3\times10^{-4}$, weight decay $10^{-4}$, and gradient clipping one. Native \texttt{verify\_MATH.grade\_answer} scores all 500 terminal answers. The complete \method{} and Plan and Budget outcomes are 477 and 474.

\subsection{Code supervision with shared initial programs}
\label{sec:matched-code}

\paragraph{Matched fitting.}
The trained arms use the same 400 earlier programming problems, 2,397 state/allowance pairs, 22-dimensional state, and two 16-wide heads. A 319/81 contest-group split supports development before the all-400 refit. AdamW uses learning rate $3\times10^{-4}$, weight decay $10^{-4}$, clipping one, and 300 epochs. Initialization uses seed 20261006. Positive, negative, and tied quality contrasts receive weights 32, 8, and 0.25, divided by the number of pairs for their problem. Relative targets use quality MSE, processing MSE, and direction BCE with coefficients 1, 1, and 0.25, at $\lambda=0.05$. Shuffled supervision jointly permutes each quality/processing target pair across the fitting population using seed 20261007. The absolute-utility arm fits $Y_{\spend}-0.05C_{\spend}/Z$ and $Y_{\preserve}-0.05C_{\preserve}/Z$, using the common per-pair processing normalizer $Z$, two squared losses, and the same direction loss. Inference subtracts its two outputs and applies the common capacity terms.

\paragraph{Matched execution.}
Each standalone policy receives the same 126 initial programs, exact initial payloads, public cases, repair prompt, Gemma actor, allowance candidates, 8,192-token generated bank, two-repair horizon, extractor, and final public/private scorer. The initial generation contributes 264,591 tokens and 126 request starts once to each arm. Each arm then contributes its own native scoring, controller work, and new requests. Fixed \preserve{} retains the initial program; feasible fixed \spend{} requests the largest legal repair allowance at each decision. Table~\ref{tab:matched-code} reports the complete terminal outcomes.

\begin{table}[t]
\centering\small
\caption{Code policies with shared initial programs and execution. Actor tokens include inherited initial generation and every new attempt. The absolute-utility token total is a lower bound because one timeout has unresolved usage.}
\label{tab:matched-code}
\begin{tabular}{@{}lrrr@{}}
\toprule
Policy & Correct /126 & Tokens & Starts \\
\midrule
Paired terminal & 93 & 506,398 & 173 \\
Shuffled terminal & 83 & 286,347 & 133 \\
Absolute utility & 92 & $\geq$525,529 & 178 \\
Fixed Preserve & 86 & 264,591 & 126 \\
Feasible fixed Spend & 92 & 1,025,266 & 371 \\
\bottomrule
\end{tabular}
\end{table}

\paragraph{Paired inference.}
The 39 contest groups are resampled 20,000 times using seed 20261007. Two-sided group sign flips use 100,000 draws with plus-one correction. Holm adjustment covers the shuffled and absolute supervision contrasts. \method{} has 11 wins and one loss against shuffled targets, yielding 7.94 points ($[2.40,14.07]$), raw $p=0.01973$, and adjusted $p=0.03946$. Against absolute utility it has two wins and one loss, yielding 0.79 points ($[-1.69,3.54]$), adjusted $p=1$. Fixed Preserve yields eight wins and one loss, 5.56 points ($[1.52,9.91]$), raw $p=0.03918$; fixed Spend yields four wins and three losses, 0.79 points ($[-3.73,5.60]$), raw $p=1$. The fixed-policy probabilities retain their separate descriptive role.

\method{} uses 50.61\% fewer actor tokens than fixed Spend, with a 95\% interval of $[41.68,60.01]\%$. Physical request starts fall from 371 to 173, a 53.37\% reduction with a 95\% interval of $[47.18,58.84]\%$. Both resource intervals use 20,000 paired resamples of the 39 contests with seed 20261007, recomputing one minus the ratio of complete totals for the respective metric on each draw. These controls use their shared initial programs and a paired refit.

\paragraph{Public-obligation features under separate refitting.}
Numerical task-state information improves native code completion in a separately refitted control family (Table~\ref{tab:code-obligation-input}). We compare the full input with an arm that sets its six numerical obligation features to zero during fitting and native execution. These fields encode public-case success, the public failure fraction, program syntax validity, a repair-history indicator, the allowance-to-remaining-bank ratio, and public-feedback availability. Both arms retain the same visible texts, status encoding, ledger, 2,400-parameter core, and native interface.

Each arm uses the same 400 earlier problems and 2,397 state/allowance pairs, the frozen 319/81 contest split, and a fresh all-400 refit after development selection. AdamW uses learning rate $3\times10^{-4}$, weight decay $10^{-4}$, and clipping one, with at most 600 development epochs and prices $\{0.005,0.05\}$. The selected full-input fit uses price 0.05 and 350 epochs; the zero-feature fit uses price 0.005 and 51 epochs. Native evaluation shares all 126 retained initial programs, Google Gemma~4 31B IT, temperature zero, the public-feedback repair interface, official LiveCodeBench scoring, an 8,192-token bank, allowances 512/2,048/4,096, and two repair opportunities.

\begin{table}[tbp]
\centering\small
\setlength{\tabcolsep}{4pt}
\caption{Public-obligation input study on all 126 code problems. Both rows are separately refitted, development-selected controllers. The full-input row belongs to this control family; the primary paired-terminal deployment in Table~\ref{tab:matched-code} completes 93/126.}
\label{tab:code-obligation-input}
\begin{tabularx}{\columnwidth}{@{}Xrr@{}}\toprule
Controller input & Correct /126 & Rate (\%)\\\midrule
Full public-obligation fields & 90 & 71.43\\
Six numerical fields set to zero & 82 & 65.08\\\bottomrule
\end{tabularx}
\end{table}

The full-input fit has nine wins, one loss, and 116 ties, giving a 6.35-point paired gain with 95\% interval $[1.67,11.19]$ and two-sided Holm-adjusted $p=0.03557$. The analysis uses 20,000 paired bootstrap draws and 100,000 whole-group sign flips over the 39 contests; adjustment retains the registered single quality contrast. Inference conditions on these development-selected fitted policies. The result supports the numerical public-obligation representation within this complete native refitting comparison.

\subsection{Retail quality and resource comparisons}
\label{sec:retail-resources}

The selected retail controller and Forest-of-Thought each complete 29/40 tasks. All forty outcomes are paired by native task identity; connected shared customer/order references form 32 dependence groups. The six \method{} wins and six losses yield a zero quality difference with a 95\% interval of $[-18.18,18.42]$ percentage points. Table~\ref{tab:retail-resources} reports all-retained-attempt processing across actor, customer simulator, and assertion judge. Controller encoding is recorded separately, and its elapsed time is nested in task time. The resource result concerns the adopted native Forest-of-Thought implementation.

\begin{table}[t]
\centering\footnotesize
\caption{Retail resources over forty tasks. Reduction intervals resample the same 32 dependence groups.}
\label{tab:retail-resources}
\setlength{\tabcolsep}{4pt}
\begin{tabular}{@{}lrrr@{}}
\toprule
Resource & \method{} & FoT & Reduction 95\% CI \\
\midrule
Model tokens & 3,331,003 & 98,364,608 & $[96.31,96.86]\%$ \\
Requests & 722 & 21,434 & $[96.40,96.82]\%$ \\
Service (s) & 10,421.51 & 151,898.33 & $[92.08,93.95]\%$ \\
Task time (s) & 10,748.02 & 152,362.67 & $[91.86,93.78]\%$ \\
\bottomrule
\end{tabular}
\end{table}

Matched fixed Preserve and Spend each complete 28/40 in their independently executed conversations. \method{}'s 2.50-point gains have intervals $[-10.26,15.63]$ and $[-10.26,15.79]$, respectively. The three retail quality comparisons jointly receive Holm correction and each has adjusted $p=1$. Quality bootstraps use seed 20261006; resource bootstraps use seed 20261007, both with 20,000 draws. The fixed controls describe the quality variation around this selected operating point.

\begin{table*}[!t]
\centering\footnotesize
\setlength{\tabcolsep}{4pt}
\caption{\textbf{Additional actors on BFCL memory.} Success rates (\%) use all 155 questions. Fit/select gives the paired fitting and selection examples. Actor tokens include native memory preparation and evaluation. Within each actor, the fixed policies share its memory interface with \method{}.}
\label{tab:backbone-controls}
\begin{tabular}{@{}lllrcrcrc@{}}
\toprule
& & & \multicolumn{2}{c}{\method{}} & \multicolumn{2}{c}{Fixed \preserve{}} & \multicolumn{2}{c}{Fixed \spend{}} \\
Actor & Memory & Fit/select & Rate & Tokens & Rate & Tokens & Rate & Tokens \\
\midrule
Mistral Small 3.2 24B & Key--value & 48 / 24; refit 72 & 40.6 & 5.84M & 38.1 & 5.90M & 36.8 & 5.78M \\
Mistral Small 3.2 24B & Vector & 160 / 40; refit 200 & 52.3 & 3.26M & 45.2 & 2.01M & 50.3 & 2.11M \\
Mistral Small 3.2 24B & Recursive & 160 / 40; refit 200 & 57.4 & 2.53M & 57.4 & 2.50M & 60.0 & 2.44M \\
Phi-4 14B & Key--value & 160 / 40; refit 200 & 32.3 & 12.40M & 33.5 & 12.79M & 34.2 & 14.91M \\
\bottomrule
\end{tabular}
\end{table*}

\subsection{Mathematical revision controls}
\label{sec:math-controls}

All comparisons use the complete 500 native problem identifiers and \texttt{verify\_MATH.grade\_answer}. The fixed policies share \method{}'s initial solutions, actor, bank, and repair interface. Plan and Budget and Forest-of-Thought retain their separately measured native recipes with the same Gemma actor revision. Paired problem bootstrap uses 20,000 draws and seed 20261006. Exact two-sided McNemar probabilities receive Holm adjustment across the four comparisons in Table~\ref{tab:math-controls}.

Physical-request reductions use all retained initial and revision attempts on the same 500 problems. Shared initial requests count once in each standalone policy; the physical campaign contains 1,191 unique requests. A separate single-endpoint resource comparison samples the seven whole subjects 20,000 times with seed 20261008 and recomputes $100(1-\sum C_A/\sum C_S)$. All seven subjects reduce request counts; leaving out any one subject retains reductions of 27.68--34.12\%. These estimates condition on the selected policy and observed subjects. Table~\ref{tab:math-subject-requests} gives each subject's complete request counts.

\begin{table}[t]
\centering\footnotesize
\caption{Physical model requests by MATH500 subject. \method{} and fixed \spend{} share each initial solution; counts include every retained initial and revision attempt. Savings are $100(1-C_{\mathrm{\method{}}}/C_{\mathrm{Spend}})$.}
\label{tab:math-subject-requests}
\setlength{\tabcolsep}{3pt}
\begin{tabular}{@{}lrrrr@{}}
\toprule
Subject & Tasks & \method{} & \spend{} & Saved (\%) \\
\midrule
Algebra & 124 & 144 & 248 & 41.94 \\
Counting \& Probability & 38 & 62 & 76 & 18.42 \\
Geometry & 41 & 60 & 82 & 26.83 \\
Intermediate Algebra & 97 & 157 & 194 & 19.07 \\
Number Theory & 62 & 78 & 124 & 37.10 \\
Prealgebra & 82 & 106 & 165 & 35.76 \\
Precalculus & 56 & 83 & 114 & 27.19 \\
\midrule
All subjects & 500 & 690 & 1,003 & 31.21 \\
\bottomrule
\end{tabular}
\end{table}

\begin{table}[t]
\centering\footnotesize
\caption{MATH500 quality relative to \method{}'s 477/500. Intervals and differences are percentage points; adjusted probabilities use the complete four-comparison family.}
\label{tab:math-controls}
\setlength{\tabcolsep}{4pt}
\begin{tabular}{@{}lrrrr@{}}
\toprule
Comparator & /500 & Gain & 95\% interval & Holm $p$ \\
\midrule
Fixed Preserve & 459 & 3.60 & $[2.00,5.40]$ & 0.000031 \\
Fixed Spend & 475 & 0.40 & $[-0.40,1.20]$ & 1.0 \\
Plan and Budget & 474 & 0.60 & $[-0.60,1.80]$ & 1.0 \\
Forest-of-Thought & 459 & 3.60 & $[1.60,5.60]$ & 0.00160 \\
\bottomrule
\end{tabular}
\end{table}

The Preserve comparison has 18 \method{} wins and zero losses; Spend has three wins and one loss; Plan and Budget has six wins and three losses; Forest-of-Thought has 22 wins and four losses. These complete pair counts connect each terminal-quality difference to its corresponding native comparison.

\subsection{Processing coefficient sensitivity}
\label{sec:processing-coefficient-sensitivity}
The mathematics study fixes the fitted controller and scales its processing coefficient by 0.5, 1, and 2. Every setting shares the same initial solutions and generates its own selected repairs through the native interface. A 1,024-token draft and feasible 1,536-token repair use the same 2,560-token bank. At this final repair, the future-capacity terms are zero. All 728 development problems select the coefficient before the 500 native test continuations. Request counts include inherited initial attempts once per standalone policy and every new repair attempt.

\section{Additional Actors on BFCL Memory}
\label{sec:backbone-study}

We fit separate controllers for two further actors, Mistral Small 3.2 24B (FP8 weights) and Phi-4 14B (BF16 weights), on BFCL memory. The controller architecture (2,400 parameters), the MiniLM encoder, the 8,192-token bank, the 128 and 1,024-token allowances, and the twenty-round limit are unchanged. Each actor uses its own memory instructions and tool serialization, shared by its three allowance policies. Table~\ref{tab:backbone-controls} compares \method{} with the fixed-allowance policies of the same actor on all 155 questions.

With Mistral, \method{} improves over both fixed policies on key--value memory (40.6\% against 38.1\% and 36.8\%) and vector memory (52.3\% against 45.2\% and 50.3\%). On recursive-summary memory it matches fixed \preserve{} at 57.4\% and is 2.6 points below fixed \spend{}. With Phi-4 on key--value memory, the three policies are within 1.9 points, and \method{} uses the fewest actor tokens. Every policy reaches a lower success rate with these actors than with Gemma, which leaves a narrower range in which allocation can change outcomes.

\definecolor{BoInk}{HTML}{20354D}
\definecolor{BoTeal}{HTML}{1C716C}
\definecolor{BoViolet}{HTML}{665376}
\definecolor{BoPale}{HTML}{F1F5F7}
\newcommand{\boBand}[2]{\par\medskip\noindent\colorbox{#1}{\parbox{\dimexpr\linewidth-2\fboxsep\relax}{\color{white}\sffamily\bfseries #2}}\par\smallskip}
\newcommand{\boContext}[1]{\noindent\colorbox{BoPale}{\parbox{\dimexpr\linewidth-2\fboxsep\relax}{\small #1}}\par\medskip}
\newcommand{\boCol}[2]{\par\smallskip\noindent\begin{minipage}[t]{\linewidth}\boBand{#1}{#2}\small\raggedright}
\newcommand{\boEndCol}{\end{minipage}\par\smallskip}
\newcommand{\boPromptBand}[1]{\par\ifdim\pagetotal>0pt\ifdim\dimexpr\pagegoal-\pagetotal\relax<6\baselineskip\newpage\fi\fi\boBand{BoInk}{#1}\nopagebreak[4]}
\newcommand{\boReadout}[1]{\par\smallskip\textbf{Native readout.} #1\par}

\section{Native Case Studies and Prompt Templates}
\label{app:cases-prompts}
The cases below pair PORA with the highest-quality recorded baseline in the corresponding main panel. They illustrate three uses of the same resource decision: repairing an unresolved solution, preserving a sufficient answer, and reserving a finite reasoning bank for a later execution boundary. The full native comparisons quantify performance; the excerpts identify the behavior of these particular tasks.
\subsection{Spending on an unresolved mathematical solution}
\label{app:case-math-spend}
\boContext{\textbf{MATH500: \texttt{test/prealgebra/874.json} (index 240).} The task asks for $\angle PQR$ with $PT\parallel QR$ and angle labels $x$, $2x$ and $128^\circ$. The supplied diagram is represented by its original Asymptote code in the actor input.}
\boCol{BoViolet}{LATS}
The recorded search runs four candidate generations and four self-evaluations. Its selected answer is 154.
\boReadout{Native correctness $=0$; 8 model requests; 10416 processed tokens.}\boEndCol\hfill\boCol{BoTeal}{PORA: Spend}
The initial draft has competing angle interpretations and fails the native answer check. The controller selects one repair; the additional generation finishes with 116.
\par\smallskip\begingroup\footnotesize\ttfamily\raggedright\setlength{\parindent}{0pt}
exchange utility = 0.\allowbreak{}939213\par
repair allowance = 1536\par
final answer = 116\par
\par\endgroup\smallskip
\boReadout{Native correctness $=1$; 2 model requests; 3,085 processed tokens.}\boEndCol
\par\medskip
The actor receives the problem, its existing draft, and a fixed checking instruction. The controller decides whether the additional computation is valuable; the native terminal scorer verifies the returned answer. The displayed utility is the recorded controller output, and the token total includes the initial generation and selected repair.
\subsection{Preserving a sufficient answer}
\label{app:case-math-preserve}
\boContext{\textbf{MATH500: \texttt{test/precalculus/807.json} (index 0).} Convert $(0,3)$ to polar coordinates $(r,\theta)$ with $r>0$ and $0\leq\theta<2\pi$. The initial answer is $(3,\pi/2)$.}
\boCol{BoViolet}{LATS}
The recorded search returns the correct polar coordinates after its candidate/evaluation procedure.
\boReadout{Native correctness $=1$; 8 model requests; 5206 processed tokens.}\boEndCol\hfill\boCol{BoTeal}{PORA: Preserve}
The controller retains the initial answer with its existing evidence.
\par\smallskip\begingroup\footnotesize\ttfamily\raggedright\setlength{\parindent}{0pt}
exchange utility = -\allowbreak{}0.\allowbreak{}0846272\par
additional model requests = 0\par
final answer = (3,\allowbreak{} pi/\allowbreak{}2)\par
\par\endgroup\smallskip
\boReadout{Native correctness $=1$; 1 model request; 403 processed tokens.}\boEndCol
\par\medskip
This case complements the repair example: a Preserve decision retains terminal quality while saving the additional actor work. The two tasks share the same mathematical generator and checking template.

\subsection{Reserving reasoning for execution closure}
\label{app:case-appworld-bank}
\boContext{\textbf{AppWorld task \texttt{099174d\_3}.} ``Place an Amazon order as per my Amazon shopping list note in SimpleNote.'' The actor is OpenAI gpt-oss-120b. PORA has an 8,192-token trajectory bank, short/long allowances of 128/1,024, and twenty execution steps.}
\boCol{BoViolet}{Plan and Budget}
The current-state planner allocates fourteen 128-token steps and three 256-token steps. The resulting episode reaches 5/7 requirements and fails the task.
\par\smallskip\begingroup\footnotesize\ttfamily\raggedright\setlength{\parindent}{0pt}
recorded step allowances:\allowbreak{}\par
128 tokens:\allowbreak{} 14 steps\par
256 tokens:\allowbreak{}  3 steps\par
\par\endgroup\smallskip
\boReadout{Task success $=0$; native requirement score $=5/7$.}\boEndCol\hfill\boCol{BoTeal}{PORA: Preserve / Spend}
The runtime uses fifteen short and five long allowances. Early source acquisition preserves the bank; late repair and write/terminal closure receive long allowances.
\par\smallskip\begingroup\footnotesize\ttfamily\raggedright\setlength{\parindent}{0pt}
step  0:\allowbreak{} Preserve,\allowbreak{} 128\par
step  7:\allowbreak{} Preserve,\allowbreak{} 128; reads the note\par
step 16:\allowbreak{} Spend,\allowbreak{}   1024; reads cart/\allowbreak{}state\par
step 19:\allowbreak{} Spend,\allowbreak{}   1024; executes closure\par
charged bank use:\allowbreak{} 7043 /\allowbreak{} 8192\par
\par\endgroup\smallskip
\boReadout{Task success $=1$; native requirement score $=7/7$.}\boEndCol
\par\medskip
The recorded deployment combines the learned residual predictor with the event-conditioned eligibility and reserve policy. The final allowance shown here is the runtime decision after that adaptation. The task demonstrates the narrative role of a finite bank over a long episode; its outcome includes the complete executor and typed schema checks.

\section{Prompt Templates and Message Assembly}
\label{app:prompt-catalog-37}
Templates below retain their implemented wording. Braced fields denote values supplied by the recorded task or public history. The following blocks specify the message assembly and expected outputs. Historical method identifiers inside quoted templates retain the execution version's wording.
\subsection{Mathematical initial answer and repair}
PORA first sends its initial problem prompt. A Spend action appends the initial assistant draft and the fixed repair user message. Preserve returns the existing answer. Both stages use temperature zero; the recorded caps are 1,024 for the initial answer and 1,536 for repair. The paired LATS prompts below specify its generation and evaluation roles.
\boPromptBand{MATH500 initial answer}
\noindent\textbf{Role:} user.\par\nopagebreak[4]\smallskip\nopagebreak[4]
\par\smallskip\begingroup\small\ttfamily\raggedright\setlength{\parindent}{0pt}
Solve this mathematics problem.\allowbreak{} Show concise reasoning and finish with the final answer in \textbackslash{}boxed\{.\allowbreak{}.\allowbreak{}.\allowbreak{}\}.\allowbreak{}\par
\smallskip
\{problem\}\par
\par\endgroup\smallskip
\smallskip\noindent\textit{Assembly.} Initial cap 1024; temperature 0.\allowbreak{}\par\medskip
\boPromptBand{MATH500 repair}
\noindent\textbf{Role:} user.\par\nopagebreak[4]\smallskip\nopagebreak[4]
\par\smallskip\begingroup\small\ttfamily\raggedright\setlength{\parindent}{0pt}
Check this solution carefully from the problem statement,\allowbreak{} correct any mathematical error,\allowbreak{} and finish with the final answer in \textbackslash{}boxed\{.\allowbreak{}.\allowbreak{}.\allowbreak{}\}.\allowbreak{}\par
\par\endgroup\smallskip
\smallskip\noindent\textit{Assembly.} Conversation:\allowbreak{} initial user/\allowbreak{}problem,\allowbreak{} assistant/\allowbreak{}draft,\allowbreak{} then this instruction.\allowbreak{} Repair cap 1536; temperature 0.\allowbreak{}\par\medskip
\boPromptBand{LATS mathematical candidate}
\noindent\textbf{Role:} user.\par\nopagebreak[4]\smallskip\nopagebreak[4]
\par\smallskip\begingroup\small\ttfamily\raggedright\setlength{\parindent}{0pt}
\{problem\}\par
Solve carefully and put the final mathematical answer in \textbackslash{}boxed\{.\allowbreak{}.\allowbreak{}.\allowbreak{}\}.\allowbreak{}\par
\par\endgroup\smallskip
\smallskip\noindent\textit{Assembly.} Four candidates and their self-\allowbreak{}evaluations; candidate caps 2048,\allowbreak{} self-\allowbreak{}evaluation caps 512; temperatures 0.\allowbreak{}4,\allowbreak{} 0.\allowbreak{}6,\allowbreak{} 0.\allowbreak{}8 and 1.\allowbreak{}0.\allowbreak{}\par\medskip
\boPromptBand{LATS mathematical evaluation}
\noindent\textbf{Role:} user.\par\nopagebreak[4]\smallskip\nopagebreak[4]
\par\smallskip\begingroup\small\ttfamily\raggedright\setlength{\parindent}{0pt}
Evaluate this solution using the visible problem.\allowbreak{} Return JSON with score in [0,\allowbreak{}1] and reflection.\allowbreak{}\par
\{solution\}\par
\par\endgroup\smallskip
\smallskip\noindent\textit{Assembly.} The previous user message contains the problem and generation instruction.\allowbreak{}\par\medskip
\boPromptBand{Plan and Budget current-state planner}
\noindent\textbf{Role:} system.\par\nopagebreak[4]\smallskip\nopagebreak[4]
\par\smallskip\begingroup\small\ttfamily\raggedright\setlength{\parindent}{0pt}
Decompose the remaining work of an interactive assistant using only the current conversation and available tools.\allowbreak{} Return one JSON object with difficulty from 1 to 5 and 2 to 5 ordered steps.\allowbreak{} Each step has text and positive credit.\allowbreak{} Credits should sum to 100.\allowbreak{} The first step must describe the next unresolved decision.\allowbreak{} Do not solve the task,\allowbreak{} predict future user messages,\allowbreak{} or use information absent from the state.\allowbreak{}\par
\par\endgroup\smallskip
\smallskip\noindent\textit{Assembly.} The AppWorld baseline supplies its current visible messages,\allowbreak{} selected tools and remaining/\allowbreak{}used bank in the planner state.\allowbreak{}\par\medskip
\boPromptBand{Plan and Budget state request}
\noindent\textbf{Role:} user.\par\nopagebreak[4]\smallskip\nopagebreak[4]
\par\smallskip\begingroup\small\ttfamily\raggedright\setlength{\parindent}{0pt}
Current state:\allowbreak{}\par
\{state\_\allowbreak{}text\}\par
Return the plan JSON now.\allowbreak{}\par
\par\endgroup\smallskip
\subsection{A filled mathematical request}
The following user messages reproduce the original problem, diagram source and checking request for the Spend case. The assistant draft is placed between them.
\boPromptBand{Filled MATH500 repair conversation}
\noindent\textbf{Role:} messages JSON.\par\nopagebreak[4]\smallskip\nopagebreak[4]
\par\smallskip\begingroup\small\ttfamily\raggedright\setlength{\parindent}{0pt}
[\par
  \{\par
    "content":\allowbreak{} "Solve this mathematics problem.\allowbreak{} Show concise reasoning and finish with the final answer in \textbackslash{}\textbackslash{}boxed\{.\allowbreak{}.\allowbreak{}.\allowbreak{}\}.\allowbreak{}\textbackslash{}n\textbackslash{}nIn the diagram,\allowbreak{} \$PT\$ is parallel to \$QR.\allowbreak{}\$ What is the measure of \$\textbackslash{}\textbackslash{}angle PQR\$ in degrees?\textbackslash{}n\textbackslash{}n[asy]\textbackslash{}ndraw((0,\allowbreak{}0)-\allowbreak{}-\allowbreak{}(10,\allowbreak{}0),\allowbreak{}black+linewidth(1));\textbackslash{}ndraw((0,\allowbreak{}0)-\allowbreak{}-\allowbreak{}(10,\allowbreak{}0),\allowbreak{}MidArrow);\textbackslash{}ndraw((10,\allowbreak{}0)-\allowbreak{}-\allowbreak{}(20,\allowbreak{}0),\allowbreak{}black+linewidth(1));\textbackslash{}ndraw((0,\allowbreak{}0)-\allowbreak{}-\allowbreak{}(-\allowbreak{}7,\allowbreak{}10)-\allowbreak{}-\allowbreak{}(7,\allowbreak{}10)-\allowbreak{}-\allowbreak{}(10,\allowbreak{}0),\allowbreak{}black+linewidth(1));\textbackslash{}ndraw((-\allowbreak{}5,\allowbreak{}10)-\allowbreak{}-\allowbreak{}(7,\allowbreak{}10),\allowbreak{}MidArrow);\textbackslash{}nlabel(\textbackslash{}"\$x\textasciicircum{}\{\textbackslash{}\textbackslash{}circ\}\$\textbackslash{}",\allowbreak{}(-\allowbreak{}6,\allowbreak{}10),\allowbreak{}SE);\textbackslash{}nlabel(\textbackslash{}"\$2x\textasciicircum{}\{\textbackslash{}\textbackslash{}circ\}\$\textbackslash{}",\allowbreak{}(7,\allowbreak{}10),\allowbreak{}SW);\textbackslash{}nlabel(\textbackslash{}"\$128\textasciicircum{}\{\textbackslash{}\textbackslash{}circ\}\$\textbackslash{}",\allowbreak{}(10,\allowbreak{}0),\allowbreak{}NE);\textbackslash{}nlabel(\textbackslash{}"\$P\$\textbackslash{}",\allowbreak{}(-\allowbreak{}7,\allowbreak{}10),\allowbreak{}N);\textbackslash{}nlabel(\textbackslash{}"\$T\$\textbackslash{}",\allowbreak{}(7,\allowbreak{}10),\allowbreak{}N);\textbackslash{}nlabel(\textbackslash{}"\$R\$\textbackslash{}",\allowbreak{}(10,\allowbreak{}0),\allowbreak{}S);\textbackslash{}nlabel(\textbackslash{}"\$Q\$\textbackslash{}",\allowbreak{}(0,\allowbreak{}0),\allowbreak{}S);\textbackslash{}n[/\allowbreak{}asy]",\allowbreak{}\par
    "role":\allowbreak{} "user"\par
  \},\allowbreak{}\par
  \{\par
    "content":\allowbreak{} "Check this solution carefully from the problem statement,\allowbreak{} correct any mathematical error,\allowbreak{} and finish with the final answer in \textbackslash{}\textbackslash{}boxed\{.\allowbreak{}.\allowbreak{}.\allowbreak{}\}.\allowbreak{}",\allowbreak{}\par
    "role":\allowbreak{} "user"\par
  \}\par
]\par
\par\endgroup\smallskip
\smallskip\noindent\textit{Assembly.} Actual request for test/\allowbreak{}prealgebra/\allowbreak{}874.\allowbreak{}json; the complete payload is retained with the case evidence.\allowbreak{}\par\medskip
\subsection{AppWorld API evidence and planning}
The API selector receives \texttt{Task:} followed by the user instruction and \texttt{Catalog:} followed by official API descriptions. Its selected identifiers populate \texttt{api\_evidence}; the plan compiler receives the available \texttt{apps}, that evidence and the trusted user instruction. Its generated plan is retained in the actor\textquotesingle{}s \texttt{plan\_context}.
\boPromptBand{Api Selector Sys}
\noindent\textbf{Role:} source template.\par\nopagebreak[4]\smallskip\nopagebreak[4]
\par\smallskip\begingroup\small\ttfamily\raggedright\setlength{\parindent}{0pt}
You select task-\allowbreak{}relevant AppWorld APIs from the official API name and description catalog.\allowbreak{}\par
Return at most 18 exact app\_\allowbreak{}name.\allowbreak{}api\_\allowbreak{}name identifiers,\allowbreak{} one per line.\allowbreak{} Cover every information source,\allowbreak{} set filter,\allowbreak{}\par
identity relation,\allowbreak{} temporal anchor,\allowbreak{} state change,\allowbreak{} notification,\allowbreak{} authentication,\allowbreak{} and completion obligation in the\par
task.\allowbreak{} First decompose the instruction into one retrieval query per noun phrase,\allowbreak{} restrictive clause,\allowbreak{} negation,\allowbreak{}\par
comparison,\allowbreak{} and grouped superlative.\allowbreak{} Select at least one exact source API for every operand before selecting write\par
APIs.\allowbreak{} A restrictive predicate must retrieve records of the entity it modifies:\allowbreak{} liked songs and liked playlists are\par
different evidence sources.\allowbreak{} In the construction songs from my playlists that I have liked,\allowbreak{} liked constrains the\par
action target songs; the construction songs from my liked playlists constrains playlists.\allowbreak{} Relative time must be\par
grounded in the AppWorld clock.\allowbreak{} A Venmo money transfer requires the direct funding-\allowbreak{}source evidence supported by\par
create\_\allowbreak{}transaction.\allowbreak{} Also preserve producer tools that can establish a missing balance or payment-\allowbreak{}card precondition\par
after a grounded execution error.\allowbreak{} A condition that contacts do not have a Venmo account requires a Venmo user\par
search in addition to phone contacts.\allowbreak{} A phrase such as songs from my song library that I have liked requires both\par
the song library and liked-\allowbreak{}song sources.\allowbreak{} A grouped superlative such as the most-\allowbreak{}played song from each playlist\par
requires playlist membership and the song detail source containing play\_\allowbreak{}count.\allowbreak{} Prefer recall when uncertain.\allowbreak{}\par
When a playlist must cover a time interval,\allowbreak{} select playlist membership and the song detail source containing each\par
song's duration.\allowbreak{} The plan must sum observed durations and compare them with the exact task-\allowbreak{}grounded interval.\allowbreak{}\par
When an email says a bill or receipt is attached,\allowbreak{} select gmail.\allowbreak{}show\_\allowbreak{}email,\allowbreak{} gmail.\allowbreak{}download\_\allowbreak{}attachment,\allowbreak{}\par
file\_\allowbreak{}system.\allowbreak{}login,\allowbreak{} and file\_\allowbreak{}system.\allowbreak{}show\_\allowbreak{}file in addition to the email listing API.\allowbreak{}\par
Do not invent identifiers and do not add explanations.\allowbreak{}\par
\par\endgroup\smallskip
\boPromptBand{Evidence Plan Sys}
\noindent\textbf{Role:} source template.\par\nopagebreak[4]\smallskip\nopagebreak[4]
\par\smallskip\begingroup\small\ttfamily\raggedright\setlength{\parindent}{0pt}
You are the evidence-\allowbreak{}dependency compiler for an AppWorld coding agent.\allowbreak{} Build a compact,\allowbreak{}\par
source-\allowbreak{}grounded execution graph for the task using the available apps:\allowbreak{} \{apps\}.\allowbreak{}\par
\smallskip
The following task-\allowbreak{}conditioned API evidence comes from the official AppWorld documentation.\allowbreak{} Use its exact names,\allowbreak{}\par
parameters,\allowbreak{} and meanings.\allowbreak{} If a required source is absent,\allowbreak{} schedule ApiDocs discovery before depending on it.\allowbreak{}\par
\smallskip
\{api\_\allowbreak{}evidence\}\par
\smallskip
Use these sections:\allowbreak{}\par
1.\allowbreak{} Semantic binding graph:\allowbreak{} first produce a span-\allowbreak{}coverage table.\allowbreak{} Give every noun phrase,\allowbreak{} restrictive clause,\allowbreak{}\par
negation,\allowbreak{} comparison,\allowbreak{} quantifier,\allowbreak{} and grouped superlative exactly one typed operator and source obligation.\allowbreak{} Then\par
identify the target entity,\allowbreak{} each restrictive predicate and the entity it modifies,\allowbreak{}\par
source collections,\allowbreak{} identity relations,\allowbreak{} comparisons,\allowbreak{} temporal phrases,\allowbreak{} and write arguments.\allowbreak{} Preserve grammatical\par
attachment.\allowbreak{} In a phrase such as songs from playlists that I liked,\allowbreak{} determine whether liked modifies songs or\par
playlists from the full task meaning,\allowbreak{} then retrieve evidence for that exact entity type.\allowbreak{} Never substitute a related\par
collection.\allowbreak{} Treat artist identity,\allowbreak{} song identity,\allowbreak{} playlist membership,\allowbreak{} liked membership,\allowbreak{} and library membership as\par
separate typed relations.\allowbreak{} When both high and low attachment remain possible,\allowbreak{} enumerate both graphs and prefer the\par
one where the restrictive predicate qualifies the final action target.\allowbreak{} Songs from my playlists that I have liked\par
requires playlist-\allowbreak{}library membership intersected with liked-\allowbreak{}song membership.\allowbreak{} Songs from my liked playlists uses\par
liked-\allowbreak{}playlist membership.\allowbreak{} When the task asks for a new object that is the same as a source object except for an\par
explicit change,\allowbreak{} add a copy-\allowbreak{}except operator.\allowbreak{} Read every source field that the creation API can preserve,\allowbreak{} inherit\par
all of them by default,\allowbreak{} and override only the explicitly changed field.\allowbreak{} This includes metadata such as pinned,\allowbreak{}\par
tags,\allowbreak{} visibility,\allowbreak{} and status in addition to title and content.\allowbreak{} Prefer structured fields such as artist\_\allowbreak{}id over\par
broad text-\allowbreak{}query relevance.\allowbreak{}\par
2.\allowbreak{} Terminal obligations:\allowbreak{} every answer field,\allowbreak{} state change,\allowbreak{} notification,\allowbreak{} and completion call required by the task.\allowbreak{}\par
   If the task is an imperative that only requests state changes,\allowbreak{} complete with answer=None.\allowbreak{} Return text only when\par
   the task explicitly asks a question or requests information.\allowbreak{} Never add a work summary as the completion answer.\allowbreak{}\par
3.\allowbreak{} Evidence ledger:\allowbreak{} for every value used by an obligation,\allowbreak{} name the exact source app,\allowbreak{} record field,\allowbreak{} and observation\par
   that must\par
   establish it.\allowbreak{} If the exact API semantics have not been observed,\allowbreak{} mark the source unresolved and put API\par
   discovery before all dependent reads.\allowbreak{} Do not invent endpoint names or substitute related concepts such as\par
   liked,\allowbreak{} saved,\allowbreak{} and library.\allowbreak{} Compare candidate API names and official descriptions against the task wording,\allowbreak{}\par
   then select the source with matching semantics.\allowbreak{} Never guess a number,\allowbreak{} identity,\allowbreak{} or membership.\allowbreak{} For long API\par
   returns,\allowbreak{} inspect the full value inside the sandbox and print only compact task-\allowbreak{}relevant fields so the decisive\par
   evidence remains visible in the next observation.\allowbreak{}\par
   For a value drawn from a conversation,\allowbreak{} preserve counterpart,\allowbreak{} direction,\allowbreak{} timestamp,\allowbreak{} and message order.\allowbreak{} When\par
   multiple numbers occur,\allowbreak{} first identify the message about the task event,\allowbreak{} then pair it with the counterpart's\par
   immediately following reply.\allowbreak{} Bind the value from that answer.\allowbreak{} Recency alone cannot select the value.\allowbreak{} A newer\par
   number about another event must remain a separate candidate and be rejected by topic and reply relation.\allowbreak{}\par
   Keyword search may locate the anchor turn,\allowbreak{} but it cannot retrieve an answer that omits the keyword.\allowbreak{} After locating\par
   the anchor,\allowbreak{} fetch the complete counterpart thread with no query,\allowbreak{} paginate it,\allowbreak{} sort it chronologically,\allowbreak{} and build\par
   directed edges between each turn and the next turn from the other speaker.\allowbreak{} Reformulate each required value as a\par
   focused question over this speaker-\allowbreak{}aware thread.\allowbreak{} Extract the answer span only from a reply connected to the\par
   task-\allowbreak{}event anchor.\allowbreak{}\par
   For phone.\allowbreak{}search\_\allowbreak{}text\_\allowbreak{}messages,\allowbreak{} use the documented parameters page\_\allowbreak{}index and page\_\allowbreak{}limit.\allowbreak{} Start with\par
   page\_\allowbreak{}index=0 and page\_\allowbreak{}limit=20.\allowbreak{} The parameter page does not exist.\allowbreak{} Continue until a short or empty page proves\par
   coverage before extracting any slot value.\allowbreak{}\par
4.\allowbreak{} Set constraints and candidate proofs:\allowbreak{} translate inclusion,\allowbreak{} exclusion,\allowbreak{} and negative conditions into explicit set\par
   operations.\allowbreak{} Require observations for every operand before computing an intersection or difference.\allowbreak{} Define the\par
   proof columns each candidate must satisfy before it can become a write argument.\allowbreak{} Translate contacts who do not\par
   have an account as relationship\_\allowbreak{}contacts minus exact\_\allowbreak{}account\_\allowbreak{}matches.\allowbreak{} Search results prove account membership\par
   only when the returned email or phone exactly equals the contact identity.\allowbreak{} An unrelated fuzzy result is not a\par
   match.\allowbreak{} Translate songs from my song library that I have liked as song\_\allowbreak{}library intersect liked\_\allowbreak{}songs.\allowbreak{} For a\par
   superlative from each group,\allowbreak{} enumerate every group,\allowbreak{} retrieve the comparison field for every member,\allowbreak{} and apply\par
   argmax independently within each group before any write.\allowbreak{}\par
   For capacity-\allowbreak{}limited migration from an Inbox into a named current-\allowbreak{}day project,\allowbreak{} left overs from the current day\par
   are the incomplete tasks already present in that target project.\allowbreak{} Sum their durations and subtract them from the\par
   schedule capacity before selecting any Inbox task.\allowbreak{} Inbox due dates do not replace this target-\allowbreak{}project source.\allowbreak{}\par
5.\allowbreak{} Temporal and identity anchors:\allowbreak{} resolve relative dates with phone.\allowbreak{}get\_\allowbreak{}current\_\allowbreak{}date\_\allowbreak{}and\_\allowbreak{}time from AppWorld.\allowbreak{} Never\par
   use the host clock,\allowbreak{} Python datetime.\allowbreak{}now,\allowbreak{} token expiry,\allowbreak{} or the newest observed record as the current date.\allowbreak{} Resolve\par
   roommate,\allowbreak{} friend,\allowbreak{} and family relations from the relationship field returned by phone.\allowbreak{}search\_\allowbreak{}contacts.\allowbreak{} A text\par
   query alone does not prove a relationship.\allowbreak{} Keep every restrictive phrase executable.\allowbreak{} Morning,\allowbreak{} afternoon,\allowbreak{} and\par
   evening must filter record timestamps.\allowbreak{} When a date contains multiple events,\allowbreak{} cluster records by time,\allowbreak{}\par
   description,\allowbreak{} and amount,\allowbreak{} then retain only the cluster satisfying every task predicate.\allowbreak{} Relationship and\par
   time-\allowbreak{}of-\allowbreak{}day phrases can never be marked descriptive or discarded.\allowbreak{}\par
6.\allowbreak{} Coverage closure:\allowbreak{} when the task says all,\allowbreak{} every,\allowbreak{} across,\allowbreak{} top,\allowbreak{} or otherwise ranges over a collection,\allowbreak{} inspect the\par
   pagination interface and read pages until the end.\allowbreak{} Track which required sources and pages have been covered.\allowbreak{}\par
   A collection containing exactly 5 items may be an AppWorld default page.\allowbreak{} Refetch page 0 with page\_\allowbreak{}limit=20 before\par
   using it.\allowbreak{} Retain candidates from every fetched page in one accumulator; later pages never replace earlier pages.\allowbreak{}\par
   When a group of people supplies replies,\allowbreak{} suggestions,\allowbreak{} approvals,\allowbreak{} or other evidence,\allowbreak{} create one thread obligation\par
   per group member.\allowbreak{} Paginate every member's complete thread independently until the target reply is found or a\par
   short or empty page proves the end.\allowbreak{} Merge the per-\allowbreak{}member evidence only after every obligation is closed.\allowbreak{} Evidence\par
   from one member can never discharge another member's thread obligation.\allowbreak{}\par
7.\allowbreak{} Structural scope:\allowbreak{} before counting,\allowbreak{} filtering,\allowbreak{} or aggregating text records,\allowbreak{} assign every line or block a region\par
   role:\allowbreak{} schema,\allowbreak{} title,\allowbreak{} legend,\allowbreak{} instance row,\allowbreak{} or footer.\allowbreak{} Define the target instance predicate from the task.\allowbreak{} Only\par
   instance rows satisfying that predicate may enter the aggregate.\allowbreak{} A symbol shown in a legend explains the schema;\par
   it is never itself an instance.\allowbreak{} Print the accepted rows before computing the final value.\allowbreak{} For repeated records\par
   embedded in a note or file,\allowbreak{} parse the full content inside the sandbox.\allowbreak{} If the instruction names a boolean or\par
   categorical predicate such as favorite,\allowbreak{} completed,\allowbreak{} accepted,\allowbreak{} or private,\allowbreak{} filter on that exact field before\par
   selecting any name or identifier.\allowbreak{} A raw long note cannot support a predicate-\allowbreak{}bound write.\allowbreak{}\par
   For a block-\allowbreak{}structured schedule keyed by day,\allowbreak{} keep the full content in a sandbox variable,\allowbreak{} derive the current\par
   weekday from phone.\allowbreak{}get\_\allowbreak{}current\_\allowbreak{}date\_\allowbreak{}and\_\allowbreak{}time,\allowbreak{} select the exact matching block,\allowbreak{} and parse its labeled scalar.\allowbreak{}\par
   Never infer a schedule value from a truncated printout or manually restate the block.\allowbreak{} Preserve source scalar\par
   units unless an observed API contract explicitly requires conversion.\allowbreak{} If the task names a user-\allowbreak{}owned library,\allowbreak{}\par
   candidate search remains inside that library; an empty or apparently infeasible result triggers source and\par
   parsing repair,\allowbreak{} not expansion to public records.\allowbreak{}\par
   Treat a literal output template as a field schema.\allowbreak{} Preserve separators and the exact width and case of every\par
   placeholder.\allowbreak{} MM/\allowbreak{}YY uses a two-\allowbreak{}digit month and two-\allowbreak{}digit year.\allowbreak{} MM/\allowbreak{}YYYY uses a four-\allowbreak{}digit year.\allowbreak{} yyyy-\allowbreak{}mm-\allowbreak{}dd keeps\par
   lowercase labels and hyphens.\allowbreak{} Instantiate one exemplar and compare its shape with the instruction before any\par
   write.\allowbreak{} Never normalize or expand a requested template.\allowbreak{}\par
   For Amazon records that expose product\_\allowbreak{}id,\allowbreak{} treat a task phrase naming a product category as a typed predicate.\allowbreak{}\par
   Call amazon.\allowbreak{}show\_\allowbreak{}product for each candidate and decide membership from the structured product\_\allowbreak{}type.\allowbreak{} Product names\par
   provide supporting evidence only and cannot close a category predicate by substring matching alone.\allowbreak{}\par
8.\allowbreak{} Dependency order:\allowbreak{} list read,\allowbreak{} transform,\allowbreak{} and write nodes.\allowbreak{} A write node may run only after all of its input evidence\par
   is observed.\allowbreak{} Inspect optional write parameters when their defaults can trigger a precondition such as a funding\par
   source.\allowbreak{} Homogeneous writes over a fully certified candidate set may share one compact loop that prints one\par
   receipt per candidate.\allowbreak{} Compile independent reads into obligation packets.\allowbreak{} Authenticate all planned apps in one\par
   early block using username types from the supplied API evidence.\allowbreak{} Fetch repeated records with one loop and print\par
   only proof columns.\allowbreak{} Do not re-\allowbreak{}query documentation already present in this plan,\allowbreak{} restate a known list in a new\par
   turn,\allowbreak{} or recount an observed collection.\allowbreak{} After all inputs close,\allowbreak{} execute homogeneous writes in one loop and\par
   complete in the same block after successful receipts.\allowbreak{} When a relationship determines a target group,\allowbreak{} retrieve\par
   exact relationship contacts first.\allowbreak{} Fetch groups from page\_\allowbreak{}index=0 with page\_\allowbreak{}limit=20 and select the group whose\par
   observed members contain the user and every relevant contact.\allowbreak{} Never infer group identity from its name,\allowbreak{} return\par
   order,\allowbreak{} or a partial member match.\allowbreak{}\par
   For venmo.\allowbreak{}create\_\allowbreak{}transaction and venmo.\allowbreak{}approve\_\allowbreak{}payment\_\allowbreak{}request,\allowbreak{} first read the balance and registered payment\par
   cards.\allowbreak{} Use the AppWorld clock to reject expired cards.\allowbreak{} Prefer a valid registered payment\_\allowbreak{}card\_\allowbreak{}id when the target\par
   write accepts it.\allowbreak{} If a newly observed supervisor card must be registered,\allowbreak{} bind the returned payment\_\allowbreak{}card\_\allowbreak{}id\par
   directly to the original target write.\allowbreak{} Do not add money to the Venmo balance when the original target write\par
   accepts payment\_\allowbreak{}card\_\allowbreak{}id.\allowbreak{} Use add\_\allowbreak{}to\_\allowbreak{}venmo\_\allowbreak{}balance only when the original target write has no direct funding-\allowbreak{}source\par
   parameter and its observed balance precondition is unmet.\allowbreak{} If no registered source can satisfy the amount,\allowbreak{} leave\par
   a runtime repair node unresolved.\allowbreak{}\par
   For file moves or renames,\allowbreak{} freeze the candidate set from the initial task-\allowbreak{}specified source directory.\allowbreak{} A\par
   destination directory is only a write target and verification source.\allowbreak{} Never modify a pre-\allowbreak{}existing destination\par
   object that was absent from the initial source collection.\allowbreak{}\par
   Before amazon.\allowbreak{}place\_\allowbreak{}order,\allowbreak{} construct the exact target product\_\allowbreak{}id-\allowbreak{}to-\allowbreak{}quantity map from task evidence.\allowbreak{} Read the\par
   current cart.\allowbreak{} For a category-\allowbreak{}scoped subset such as all weightlifting benches,\allowbreak{} call amazon.\allowbreak{}show\_\allowbreak{}product for every\par
   cart product and classify it by the structured product\_\allowbreak{}type; a name substring is only supporting evidence.\allowbreak{}\par
   Delete every unrelated product,\allowbreak{} add every missing product,\allowbreak{} and correct every mismatched quantity.\allowbreak{}\par
   Read the cart again and require exact equality with the target map before placing the order.\allowbreak{} Existing cart state\par
   is untrusted task context; never allow an unrelated product to enter the new order.\allowbreak{}\par
   A cart mutation and amazon.\allowbreak{}place\_\allowbreak{}order must occur in separate agent turns.\allowbreak{} After the final cart mutation,\allowbreak{} end\par
   that turn with amazon.\allowbreak{}show\_\allowbreak{}cart.\allowbreak{} Inspect its next observation,\allowbreak{} print the exact observed map,\allowbreak{} and only then place\par
   the order in a later turn.\allowbreak{} A cart observation produced earlier in the same code block cannot certify the write.\allowbreak{}\par
   Treat a wish list as a read-\allowbreak{}only candidate source for a single-\allowbreak{}product purchase.\allowbreak{} Add the selected product to the\par
   cart while preserving its wish-\allowbreak{}list entry.\allowbreak{} When the instruction quantifies over the wish list itself,\allowbreak{} such as\par
   buying all items or as many items as possible,\allowbreak{} move every selected entry from the wish list to the cart so the\par
   collection-\allowbreak{}level state transition is discharged.\allowbreak{}\par
   Read registered Amazon payment cards and addresses before ordering.\allowbreak{} Select an observed unexpired card using the\par
   AppWorld clock and an observed address.\allowbreak{} If a named address is absent from Amazon,\allowbreak{} read the exact user-\allowbreak{}owned\par
   address from supervisor,\allowbreak{} register that observed address in Amazon,\allowbreak{} and bind the returned address\_\allowbreak{}id.\allowbreak{} If no\par
   registered Amazon card can satisfy the purchase,\allowbreak{} read an observed user-\allowbreak{}owned card from supervisor,\allowbreak{} register it\par
   in Amazon,\allowbreak{} and bind the returned payment\_\allowbreak{}card\_\allowbreak{}id.\allowbreak{} Never guess identifiers or synthesize field values.\allowbreak{}\par
9.\allowbreak{} Runtime repair:\allowbreak{} an execution failure is a new observation,\allowbreak{} not permission to guess.\allowbreak{} Convert its exact unmet\par
   condition into a typed deficit.\allowbreak{} Search backward for the shortest producer chain whose postcondition satisfies\par
   that deficit.\allowbreak{} Every producer input must come from a printed,\allowbreak{} user-\allowbreak{}owned source.\allowbreak{} A cross-\allowbreak{}app repair is admissible\par
   only when the failing target action explicitly requires that resource and the repair changes no unrelated user\par
   intent.\allowbreak{} Minimize changed model types as well as call count.\allowbreak{} When the original write accepts a direct source\par
   parameter,\allowbreak{} a support action may register that source,\allowbreak{} but it must not create an intermediate balance,\allowbreak{} transfer,\allowbreak{}\par
   message,\allowbreak{} file,\allowbreak{} or notification.\allowbreak{} Execute one verified repair path,\allowbreak{} print its receipt,\allowbreak{} verify the new state,\allowbreak{} and\par
   retry the original write once.\allowbreak{} If no path is certified,\allowbreak{} continue discovery or stop without speculative identifiers.\allowbreak{}\par
10.\allowbreak{} Closure checks:\allowbreak{} immediately before each write,\allowbreak{} state the exact arguments and their observed provenance.\allowbreak{} At the\par
   end,\allowbreak{} verify every terminal obligation and then call supervisor.\allowbreak{}complete\_\allowbreak{}task(answer=\textless{}answer or None\textgreater{}).\allowbreak{}\par
\smallskip
Include API discovery whenever the exact interface is unknown.\allowbreak{} Name variables that later steps will need.\allowbreak{} Respond\par
only with the structured plan.\allowbreak{}\par
\par\endgroup\smallskip
\subsection{AppWorld code actor}
The first user message is \texttt{Task:} followed by the instruction. Each executed Python block is appended as an assistant message; the next user message is \texttt{Output:} followed by the normalized observation (up to 2,500 characters). The runtime supplies the Preserve/Spend allowance through decoding. The output is one executable Python block per step.
\boPromptBand{Sys}
\noindent\textbf{Role:} source template.\par\nopagebreak[4]\smallskip\nopagebreak[4]
\par\smallskip\begingroup\small\ttfamily\raggedright\setlength{\parindent}{0pt}
You are a coding agent solving a task for your supervisor in AppWorld.\allowbreak{} You act by writing Python code.\allowbreak{}\par
\smallskip
Rules:\allowbreak{}\par
-\allowbreak{} Respond with EXACTLY ONE ```python .\allowbreak{}.\allowbreak{}.\allowbreak{} ``` code block per turn.\allowbreak{} It runs in a STATEFUL python sandbox where a\par
  global object `apis` is available.\allowbreak{} You see its stdout as the next message.\allowbreak{} PRINT anything you want to observe.\allowbreak{}\par
-\allowbreak{} Available apps:\allowbreak{} \{apps\}.\allowbreak{} Learn an app's APIs with:\allowbreak{}\par
    print(apis.\allowbreak{}api\_\allowbreak{}docs.\allowbreak{}show\_\allowbreak{}api\_\allowbreak{}descriptions(app\_\allowbreak{}name='spotify'))\par
    print(apis.\allowbreak{}api\_\allowbreak{}docs.\allowbreak{}show\_\allowbreak{}api\_\allowbreak{}doc(app\_\allowbreak{}name='spotify',\allowbreak{} api\_\allowbreak{}name='login'))\par
-\allowbreak{} Your supervisor's info \& credentials:\allowbreak{}\par
    print(apis.\allowbreak{}supervisor.\allowbreak{}show\_\allowbreak{}profile())            \# name,\allowbreak{} email,\allowbreak{} phone\_\allowbreak{}number\par
    print(apis.\allowbreak{}supervisor.\allowbreak{}show\_\allowbreak{}account\_\allowbreak{}passwords())  \# passwords to log into the apps\par
-\allowbreak{} Store the returned profile and credential records in sandbox variables.\allowbreak{} Select a credential record by its\par
  account\_\allowbreak{}name and pass its password field directly to login.\allowbreak{} Never manually retype credential strings or repair\par
  their punctuation.\allowbreak{}\par
-\allowbreak{} Most apps need login first:\allowbreak{} access\_\allowbreak{}token = apis.\allowbreak{}spotify.\allowbreak{}login(username=.\allowbreak{}.\allowbreak{}.\allowbreak{},\allowbreak{} password=.\allowbreak{}.\allowbreak{}.\allowbreak{})['access\_\allowbreak{}token'].\allowbreak{}\par
-\allowbreak{} When done,\allowbreak{} call:\allowbreak{} apis.\allowbreak{}supervisor.\allowbreak{}complete\_\allowbreak{}task(answer=\textless{}concise answer or None if not a question\textgreater{}).\allowbreak{}\par
-\allowbreak{} Work by evidence boundaries.\allowbreak{} Batch independent logins and homogeneous reads or writes in compact loops.\allowbreak{} Query\par
  ApiDocs only when the supplied plan lacks the schema or execution reports a schema error.\allowbreak{}\par
-\allowbreak{} Put required source reads and state-\allowbreak{}changing writes in separate environment turns.\allowbreak{} A block containing an actual\par
  state change before its required reads close is rejected in full.\allowbreak{} When an answer-\allowbreak{}only block places\par
  supervisor.\allowbreak{}complete\_\allowbreak{}task after unresolved source reads,\allowbreak{} the runtime may execute only the safe top-\allowbreak{}level read\par
  prefix and defer completion to the next turn.\allowbreak{} Keep the final completion call as a separate top-\allowbreak{}level statement\par
  so this recovery remains available.\allowbreak{} A task-\allowbreak{}required transition followed by observation is one closed-\allowbreak{}loop\par
  certificate:\allowbreak{} execute the transition,\allowbreak{} observe the resulting state,\allowbreak{} and test the exact stop condition before\par
  terminal completion.\allowbreak{}\par
-\allowbreak{} For a value owed as per a text conversation,\allowbreak{} identify the question matching the exact task event and bind its\par
  immediately following counterpart reply.\allowbreak{} Carry only the scalar from that reply into the write.\allowbreak{} A later unrelated\par
  amount or a generic first numeric mention cannot satisfy the event relation.\allowbreak{}\par
-\allowbreak{} Normalize AppWorld calendar dates before comparison.\allowbreak{} If phone.\allowbreak{}get\_\allowbreak{}current\_\allowbreak{}date\_\allowbreak{}and\_\allowbreak{}time returns a value such as\par
  Thursday,\allowbreak{} May 18,\allowbreak{} 2023,\allowbreak{} parse it with datetime.\allowbreak{}strptime(value,\allowbreak{} "\%A,\allowbreak{} \%B \%d,\allowbreak{} \%Y").\allowbreak{}date().\allowbreak{}isoformat() and compare\par
  that YYYY-\allowbreak{}MM-\allowbreak{}DD value with the date prefix of an ISO record timestamp.\allowbreak{} Never compare the two raw strings.\allowbreak{}\par
\{plan\_\allowbreak{}context\}\par
\par\endgroup\smallskip
\subsection{Memory, code and retail adapters}
BFCL appends the official user turns and tool observations after the displayed memory system message. LiveCodeBench sends the public problem/interface text initially and uses the displayed repair request with public execution feedback. Retail appends the official conversation and observations after its system message. The learned controller determines the additional resource allowance at each supported decision.
\boPromptBand{System}
\noindent\textbf{Role:} source template.\par\nopagebreak[4]\smallskip\nopagebreak[4]
\par\smallskip\begingroup\small\ttfamily\raggedright\setlength{\parindent}{0pt}
Operate the official BFCL-\allowbreak{}v4 memory tools.\allowbreak{} Use only supplied tools and observed results.\allowbreak{} During prerequisites,\allowbreak{} keep stable identity in seven core slots and condense all new episodic facts from one user turn into ONE archival note (at most 2000 characters) under a descriptive snake\_\allowbreak{}case key.\allowbreak{} Archive has 50 slots.\allowbreak{} Do not make one entry per fact.\allowbreak{} If add says full,\allowbreak{} do not retry.\allowbreak{} List keys; inspect two related older entries,\allowbreak{} combine their distinct facts into one concise value with replace,\allowbreak{} then remove a redundant key only after preserving its content.\allowbreak{} Add one new note.\allowbreak{} If no safe compression fits,\allowbreak{} replace a related note with a combined value rather than discarding a distinct fact.\allowbreak{} Core values must stay under 300 characters.\allowbreak{} Use at most fourteen memory calls per turn,\allowbreak{} then answer the current user.\allowbreak{} For a question,\allowbreak{} search,\allowbreak{} list keys if search scores are zero,\allowbreak{} retrieve the needed value,\allowbreak{} and give a short exact answer copied from evidence.\allowbreak{} Never invent memory contents or hidden answers.\allowbreak{}\par
\par\endgroup\smallskip
\boPromptBand{LiveCodeBench repair}
\noindent\textbf{Role:} user.\par\nopagebreak[4]\smallskip\nopagebreak[4]
\par\smallskip\begingroup\small\ttfamily\raggedright\setlength{\parindent}{0pt}
Review the current Python solution using the public cases and public execution feedback.\allowbreak{} Return one complete Python code block.\allowbreak{}\par
\{problem\}\par
Current code:\allowbreak{}\par
\{code\}\par
Public cases:\allowbreak{}\par
\{public\_\allowbreak{}cases\_\allowbreak{}json\}\par
Public execution feedback:\allowbreak{}\par
\{feedback\_\allowbreak{}json\}\par
\par\endgroup\smallskip
\smallskip\noindent\textit{Assembly.} Initial generation sends the task/\allowbreak{}interface text as one user message; repair caps are selected from 512,\allowbreak{} 2048 and 4096 within the 8192-\allowbreak{}token bank.\allowbreak{}\par\medskip
\boPromptBand{Retail actor system}
\noindent\textbf{Role:} system.\par\nopagebreak[4]\smallskip\nopagebreak[4]
\par\smallskip\begingroup\small\ttfamily\raggedright\setlength{\parindent}{0pt}
You are the customer-\allowbreak{}service agent in the official tau3\par
retail benchmark.\allowbreak{} Follow the supplied retail policy exactly.\allowbreak{} In each turn,\allowbreak{}\par
return either a user-\allowbreak{}facing text response or one or more official tool calls,\allowbreak{}\par
never both.\allowbreak{} Never invent tool observations.\allowbreak{} Tool observations are generated\par
only by the official tau2 environment.\allowbreak{}\par
\smallskip
\textless{}retail\_\allowbreak{}policy\textgreater{}\par
\{domain\_\allowbreak{}policy\}\par
\textless{}/\allowbreak{}retail\_\allowbreak{}policy\textgreater{}\par
\smallskip
\textless{}method\_\allowbreak{}controller\textgreater{}\par
Use the COT2 unified Preserve/\allowbreak{}Spend controller and keep the official Tau3 retail tool protocol.\allowbreak{}\par
\textless{}/\allowbreak{}method\_\allowbreak{}controller\textgreater{}\par
\par\endgroup\smallskip
\smallskip\noindent\textit{Assembly.} The retained executor uses the displayed historical controller identifier.\allowbreak{} PORA supplies the measured residual budget selector; the retail policy is the official task environment policy.\allowbreak{}\par\medskip

\section{Native Dialogue Simulation and Scoring}
\label{app:prompt-catalog-38}
Templates below retain their implemented wording. Braced fields denote values supplied by the recorded task or public history. The following blocks specify the message assembly and expected outputs. Historical method identifiers inside quoted templates retain the execution version's wording.
\subsection{User simulator and terminal judge}
The retail actor and controller use the public dialogue. These native auxiliary roles produce the customer continuation and terminal assertion scores. The task scenario fills instructions; persona guidelines populate the displayed marker. The measured actor, simulator and judge calls use the recorded Gemma 4 adapter and retain separate request accounting.
\boPromptBand{Native retail user simulator}
\noindent\textbf{Role:} system.\par\nopagebreak[4]\smallskip\nopagebreak[4]
\par\smallskip\begingroup\small\ttfamily\raggedright\setlength{\parindent}{0pt}
\smallskip
\{global\_\allowbreak{}user\_\allowbreak{}sim\_\allowbreak{}guidelines\_\allowbreak{}with\_\allowbreak{}persona\}\par
\smallskip
\textless{}scenario\textgreater{}\par
\{instructions\}\par
\textless{}/\allowbreak{}scenario\textgreater{}\par
\par\endgroup\smallskip
\smallskip\noindent\textit{Assembly.} The global guidelines below supply the first field,\allowbreak{} followed by the task scenario.\allowbreak{} Public agent messages drive the simulator.\allowbreak{}\par\medskip
\boPromptBand{Native retail simulation guidelines}
\noindent\textbf{Role:} system field.\par\nopagebreak[4]\smallskip\nopagebreak[4]
\par\smallskip\begingroup\small\ttfamily\raggedright\setlength{\parindent}{0pt}
\# User Simulation Guidelines\par
You are playing the role of a customer contacting a customer service representative.\allowbreak{} \par
Your goal is to simulate realistic customer interactions while following specific scenario instructions.\allowbreak{}\par
\smallskip
\#\# Core Principles\par
-\allowbreak{} Generate one message at a time,\allowbreak{} maintaining natural conversation flow.\allowbreak{}\par
-\allowbreak{} Strictly follow the scenario instructions you have received.\allowbreak{}\par
-\allowbreak{} Never make up or hallucinate information not provided in the scenario instructions.\allowbreak{} Information that is not provided in the scenario instructions should be considered unknown or unavailable.\allowbreak{}\par
-\allowbreak{} Avoid repeating the exact instructions verbatim.\allowbreak{} Use paraphrasing and natural language to convey the same information\par
-\allowbreak{} Disclose information progressively.\allowbreak{} Wait for the agent to ask for specific information before providing it.\allowbreak{}\par
\smallskip
\#\# Task Completion\par
-\allowbreak{} The goal is to continue the conversation until the task is complete.\allowbreak{}\par
-\allowbreak{} If the instruction goal is satisified,\allowbreak{} generate the '\#\#\#STOP\#\#\#' token to end the conversation.\allowbreak{}\par
-\allowbreak{} If you are transferred to another agent,\allowbreak{} generate the '\#\#\#TRANSFER\#\#\#' token to indicate the transfer.\allowbreak{}\par
-\allowbreak{} If you find yourself in a situation in which the scenario does not provide enough information for you to continue the conversation,\allowbreak{} generate the '\#\#\#OUT-\allowbreak{}OF-\allowbreak{}SCOPE\#\#\#' token to end the conversation.\allowbreak{}\par
Remember:\allowbreak{} The goal is to create realistic,\allowbreak{} natural conversations while strictly adhering to the provided instructions and maintaining character consistency.\allowbreak{}\par
\par\endgroup\smallskip
\boPromptBand{Native retail natural-language judge}
\noindent\textbf{Role:} system.\par\nopagebreak[4]\smallskip\nopagebreak[4]
\par\smallskip\begingroup\small\ttfamily\raggedright\setlength{\parindent}{0pt}
\smallskip
        TASK\par
        -\allowbreak{} You will be given a list of expected outcomes and a conversation that was collected during a test case run.\allowbreak{}\par
        -\allowbreak{} The conversation is between an agent and a customer.\allowbreak{}\par
        -\allowbreak{} Your job is to evaluate whether the agent satisfies each of the expected outcomes.\allowbreak{}\par
        -\allowbreak{} Grade each expected outcome individually.\allowbreak{}\par
\smallskip
        FORMAT\par
        -\allowbreak{} Your response should be a JSON object with the following fields:\allowbreak{}\par
        -\allowbreak{} `reasoning`:\allowbreak{} a short explanation for your classification\par
        -\allowbreak{} `metExpectation`:\allowbreak{} `true` if the agent satisfies the expected outcomes,\allowbreak{} `false` otherwise\par
        -\allowbreak{} `expectedOutcome`:\allowbreak{} repeat the expectation from the input that you are grading\par
        \par
        Example response structure:\allowbreak{}\par
        \{\par
            "results":\allowbreak{} [\par
                \{\par
                    "expectedOutcome":\allowbreak{} "\textless{}one of the expected outcomes from the input\textgreater{}",\allowbreak{}\par
                    "reasoning":\allowbreak{} "\textless{}reasoning trace\textgreater{}",\allowbreak{}\par
                    "metExpectation":\allowbreak{} \textless{}false or true\textgreater{},\allowbreak{}\par
                \}\par
            ]\par
        \}\par
        \par
\par\endgroup\smallskip
\smallskip\noindent\textit{Assembly.} A user message supplies conversation and expectedOutcomes.\allowbreak{} The native parser reads results[].\allowbreak{}metExpectation and expectedOutcome; environment-\allowbreak{}effect checks retain their separate native evaluator.\allowbreak{}\par\medskip
\boPromptBand{Native retail judge input}
\noindent\textbf{Role:} user.\par\nopagebreak[4]\smallskip\nopagebreak[4]
\par\smallskip\begingroup\small\ttfamily\raggedright\setlength{\parindent}{0pt}
conversation:\allowbreak{}\par
\{trajectory\_\allowbreak{}str\}\par
\smallskip
expectedOutcomes:\allowbreak{}\par
\{nl\_\allowbreak{}assertions\}\par
\par\endgroup\smallskip

\end{document}